\documentclass{article}

\usepackage{microtype}
\usepackage{subfigure}
\usepackage{array}
\usepackage{hyperref}
\usepackage{amsfonts}
\usepackage{xfrac}
\usepackage{url}
\usepackage{dblfloatfix}
\usepackage{pifont}
\usepackage{graphicx}
\usepackage[utf8]{inputenc}
\usepackage{multirow}
\usepackage{booktabs}
\usepackage[symbol]{footmisc}

\usepackage{hyperref}

\usepackage[accepted,nohyperref]{icml2020}

\icmltitlerunning{Rigging the Lottery: Making All Tickets Winners}

\begin{document}

\twocolumn[
\icmltitle{Rigging the Lottery: \\Making All Tickets Winners}



\icmlsetsymbol{equal}{*}

\begin{icmlauthorlist}
\icmlauthor{Utku Evci}{goo}
\icmlauthor{Trevor Gale}{goo}
\icmlauthor{Jacob Menick}{dm}
\icmlauthor{Pablo Samuel Castro}{goo}
\icmlauthor{Erich Elsen}{dm}
\end{icmlauthorlist}

\icmlaffiliation{dm}{DeepMind}
\icmlaffiliation{goo}{Google Brain}

\icmlcorrespondingauthor{Utku Evci}{evcu@google.com}
\icmlcorrespondingauthor{Erich Elsen}{eriche@google.com}

\icmlkeywords{Machine Learning, ICML}

\vskip 0.3in
]



\printAffiliationsAndNotice{}  

\begin{abstract}
Many applications require sparse neural networks due to space or inference time restrictions. 
There is a large body of work on training dense networks to yield sparse networks for inference,  
but this limits the size of the largest trainable sparse model to that of the largest trainable dense model. 
In this paper we introduce a method to train sparse neural networks with a fixed parameter count and a fixed computational cost throughout training, without sacrificing accuracy relative to existing dense-to-sparse training methods. Our method updates the topology of the sparse network during training by using parameter magnitudes and infrequent gradient calculations.  We show that this approach requires fewer floating-point operations (FLOPs) to achieve a given level of accuracy compared to prior techniques.  
We demonstrate state-of-the-art sparse training results on a variety of networks and datasets, including ResNet-50, MobileNets on Imagenet-2012, and RNNs on WikiText-103.
Finally, we provide some insights into why allowing the topology to change during the optimization can overcome local minima encountered when the topology remains static\footnote{Code available at \href{https://github.com/google-research/rigl}{github.com/google-research/rigl}}.
\end{abstract}
\section{Introduction}
\label{sec:intro}
The parameter and floating point operation (FLOP) efficiency of sparse neural networks is now well demonstrated on a variety of problems~\citep{han2015learning, pruning_spike_slab_prior}.  Multiple works have shown inference time speedups are possible using sparsity for both Recurrent Neural Networks (RNNs)~\citep{kalchbrenner2018} and Convolutional Neural Networks (ConvNets)~\citep{SparseCNN_Intel_Park16, elsen2019fast}.  Currently, the most accurate sparse models are obtained with techniques that require, at a minimum, the cost of training a dense model in terms of memory and FLOPs~\citep{gupta2018, DynamicSurgery}, and sometimes significantly more~\citep{variational-dropout}. 

\begin{figure}
\centering
  \includegraphics[width=.8\linewidth]{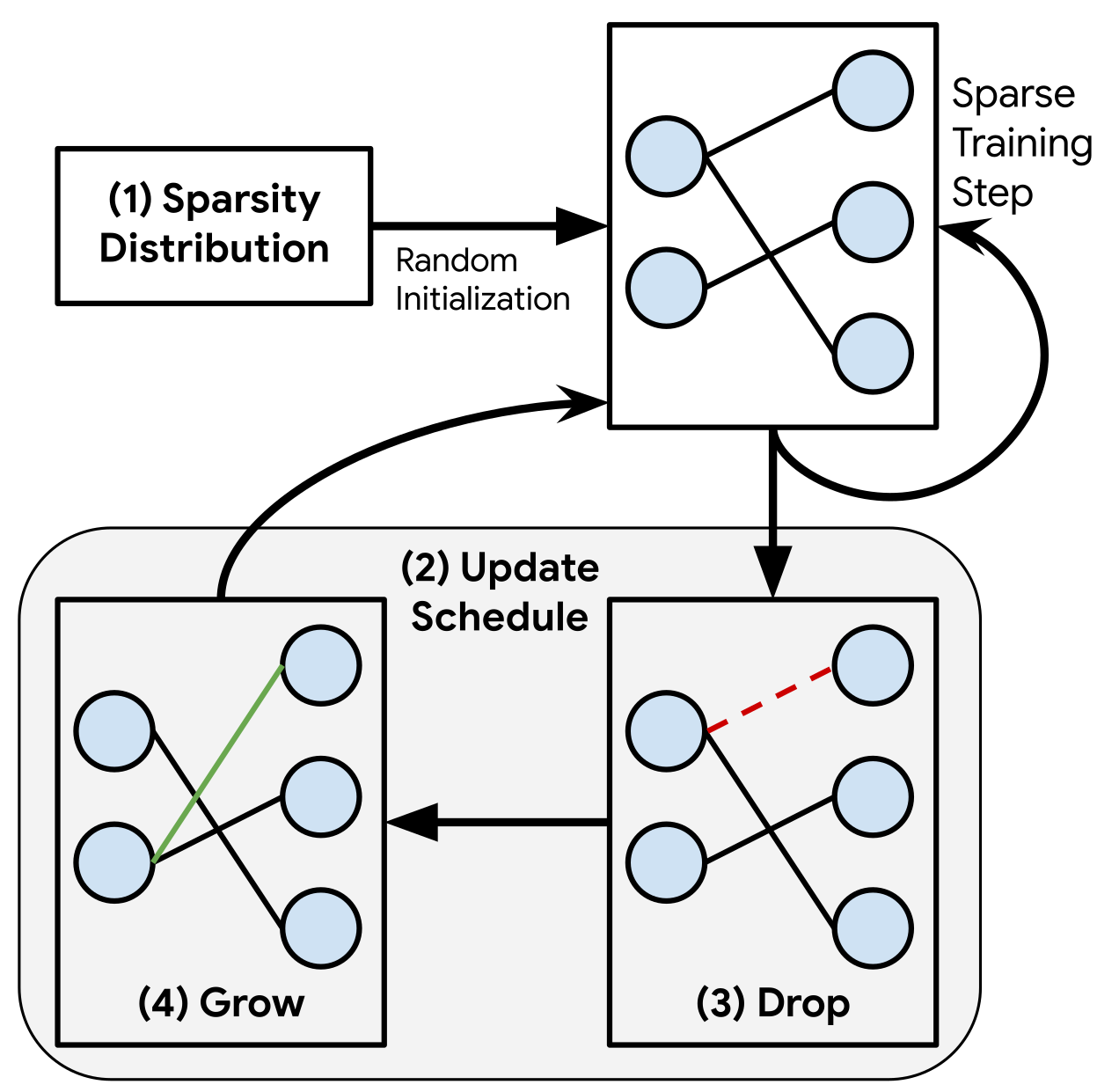}
\caption{\textit{RigL} improves the optimization of sparse neural networks by leveraging weight magnitude and gradient information to jointly optimize model parameters and connectivity.}
\label{fig:strain}
\end{figure}

This paradigm has two main limitations.  First, the maximum size of sparse models is limited to the largest dense model that can be trained;  even if sparse models are more parameter efficient, we can't use pruning to train models that are larger and more accurate than the largest possible dense models.  Second, it is inefficient; large amounts of computation must be performed for parameters that are zero valued or that will be zero during inference. Additionally, it remains unknown if the performance of the current best pruning algorithms is an upper bound on the quality of sparse models.  \citet{gale2019state} found that three different dense-to-sparse training algorithms all achieve about the same sparsity / accuracy trade-off. However, this is far from conclusive proof that no better performance is possible.  


The Lottery Ticket Hypothesis~\citep{frankle2018} hypothesized that if we can find a sparse neural network with iterative pruning, then we can train that sparse network from scratch, to the same level of accuracy, by {\em starting from the original initial conditions}. In this paper we introduce a new method for training sparse models without the need of a ``lucky'' initialization; for this reason, we call our method ``The Rigged Lottery'' or \textit{RigL}\footnote{Pronounced "wriggle".}. We make the following specific contributions:

\begin{itemize}
  \setlength\itemsep{.1em}
  \item  We introduce \textit{RigL} - an algorithm for training sparse neural networks while maintaining memory and computational cost proportional to density of the network.
  \item We perform an extensive empirical evaluation of \textit{RigL} on computer vision and natural language tasks. We show that \textit{RigL} achieves higher quality than all previous techniques for a given computational cost. 
  \item We show the surprising result that \textit{RigL} can find more accurate models than the current best dense-to-sparse training algorithms. 
  \item We study the loss landscape of sparse neural networks and provide insight into why allowing the topology of nonzero weights to change over the course of training aids optimization.
\end{itemize}




\begin{table*}
    \centering
    \begin{tabular}{c|c|c|c|c}
        Method & Drop & Grow & Selectable FLOPs & Space \& FLOPs $\propto$ \\\hline
        SNIP & $min(|\theta * \nabla_{\theta} L(\theta)|)$ & none & yes & sparse \\
        DeepR & stochastic& random & yes & sparse\\
        SET  & $min(|\theta|)$ & random & yes & sparse  \\
        DSR  & $min(|\theta|)$ & random & no & sparse \\
        SNFS & $min(|\theta|)$ & momentum & no & dense \\
        RigL (ours) & $min(|\theta|)$ & gradient & yes & sparse \\
    \end{tabular}
    \caption{Comparison of different sparse training techniques. \textit{Drop} and \textit{Grow}
    columns correspond to the strategies used during the mask update. \textit{Selectable FLOPs} is possible if the cost of training the model is fixed at the beginning of training.}
    \label{tab:methods}
\end{table*}

\section{Related Work}

\label{sec:related}
Research on finding sparse neural networks dates back decades, at least to \citet{Thimm95evaluatingpruning} who concluded that pruning weights based on magnitude was a simple and powerful technique.  \citet{sparse-connection-1997} later introduced the idea of retraining the previously pruned network to increase accuracy. \citet{deepcompression} went further and introduced multiple rounds of magnitude pruning and retraining.  This is, however, relatively inefficient, requiring ten rounds of retraining when removing $20\%$ of the connections to reach a final sparsity of $90\%$.  To overcome this problem, \citet{exploring-sparsity-rnn} introduced gradual pruning, where connections are slowly removed over the course of a single round of training.  \citet{gupta2018} refined the technique to minimize the amount of hyper-parameter selection required.

A diversity of approaches not based on magnitude pruning have also been proposed.  \citet{mozer1989}, \citet{lecun1990} and \citet{hassibi1993} are some early examples, but impractical for modern neural networks as they use information from the Hessian to prune a trained network.  More recent work includes $L_0$ Regularization~\citep{Louizos2018}, Variational Dropout~\citep{variational-dropout}, Dynamic Network Surgery~\citep{DynamicSurgery}, Discovering Neural Wirings~\citep{wortsman2019dnw}, Sensitivity Driven Regularization~\citep{SparsityDrivenRegularization}. \citet{gale2019state} examined magnitude pruning, $L_0$ Regularization, and Variational Dropout and concluded that they all achieve about the same accuracy versus sparsity trade-off on ResNet-50 and Transformer architectures.

There are also structured pruning methods which attempt to remove channels or neurons so that the resulting network is dense and can be accelerated easily \citep{vib2018, sbp2017, Louizos2018}. We compare \textit{RigL} with these state-of-the-art structured pruning methods in Appendix ~\ref{app:mnist}. We show that our method requires far fewer resources and finds smaller networks that require less FLOPs to run.

The first training technique to allow for sparsity throughout the entire training process was, to our knowledge, first introduced in Sparse Evolutionary Training (SET)~\citep{Mocanu2018}.  The weights are pruned according to the standard magnitude criterion used in pruning and are added back at random.  The method is simple and achieves reasonable performance in practice. Following this, Deep Rewiring (DeepR)~\citep{Bellec2017} augmented Stochastic Gradient Descent (SGD) with a random walk in parameter space.  Additionally, at initialization connections are assigned a pre-defined sign at random; when the optimizer would normally flip the sign, the weight is set to 0 instead and new weights are activated at random. 



Dynamic Sparse Reparameterization (DSR)~\citep{Mostafa2019} introduced the idea of allowing the parameter budget to shift between different layers of the model, allowing for non-uniform sparsity. This allows the model to distribute parameters where they are most effective. Unfortunately, the models under consideration are mostly convolutional networks, so the result of this parameter reallocation (which is to decrease the sparsity of early layers and increase the sparsity of later layers) has the overall effect of increasing the FLOP count because the spatial size is largest in the early layers. Sparse Networks from Scratch (SNFS)~\citep{dettmers2019} introduces the idea of using the momentum of each parameter as the criterion to be used for growing weights and demonstrates it leads to an improvement in test accuracy. Like DSR, they allow the sparsity of each layer to change and focus on a constant parameter, not FLOP, budget. Importantly, the method requires computing gradients and updating the momentum for \emph{every} parameter in the model, even those that are zero, at \emph{every} iteration. This can result in a significant amount of overall computation. Additionally, depending on the model and training setup, the required storage for the full momentum tensor could be prohibitive. Single-Shot Network Pruning (SNIP)~\citep{SNIP} attempts to find an initial mask with one-shot pruning and uses the saliency score of parameters to decide which parameters to keep. After pruning, training proceeds with this static sparse network. Properties of the different sparse training techniques are summarized in \autoref{tab:methods}.

There has also been a line of work investigating the Lottery Ticket Hypothesis \citep{frankle2018}. \citet{frankle2019} showed that the formulation must be weakened to apply to larger networks such as ResNet-50 \citep{He2015}. In large networks, instead of the original initialization, the values after thousands of optimization steps must be used for initialization. \cite{deconstructing_lottery} showed that "winning lottery tickets" obtain non-random accuracies even before training has started. Though the possibility of training sparse neural networks with a fixed sparsity mask using lottery tickets is intriguing, it remains unclear whether it is possible to generate such initializations -- for both masks and parameters -- \emph{de novo}.

\section{Rigging The Lottery}
\label{sec:rlottery}
Our method, \textit{RigL}, is illustrated in Figure \ref{fig:strain} and detailed in Algorithm~\ref{alg:rigl}. {\em RigL} starts with a random sparse network, and at regularly spaced intervals it removes a fraction of connections based on their magnitudes and activates new ones using instantaneous gradient information. After updating the connectivity, training continues with the updated network until the next update. The main parts of our algorithm, \textit{Sparsity Distribution}, \textit{Update Schedule}, \textit{Drop Criterion}, \textit{Grow Criterion}, and the various options considered for each, are explained below.

\textbf{(0) Notation.} Given a dataset $D$ with individual samples $x_i$ and targets $y_i$, we aim to minimize the loss function $\sum_i L(f_{\Theta}(x_i), y_i)$, where $f_{\Theta}(\cdot)$ is a neural network with parameters $\Theta \in \mathbb{R}^N$. Parameters of the $l^{th}$ layer are denoted with $\Theta^l$ which is a length $N^l$ vector. A sparse layer drops a fraction $s^l \in (0,1)$ of its connections and  parameterized with vector $\theta^l$ of length $(1-s^l)N^l$. Parameters of the corresponding sparse network is denoted with $\theta$.  Finally, the overall sparsity of a sparse network is defined as the ratio of zeros to the total parameter count, i.e. $S=\frac{\sum_l s^lN^l}{N}$  

\textbf{(1) Sparsity Distribution.} There are many ways of distributing the non-zero weights across the layers while maintaining a certain overall sparsity. We avoid re-allocating parameters between layers during the training process as it makes it difficult to target a specific final FLOP budget, which is important for many applications. We consider the following three strategies:

\begin{enumerate}
    \item \textit{Uniform:} The sparsity $s^l$ of each individual layer is equal to the total sparsity $S$. In this setting, we keep the first layer dense, since sparsifying this layer has a disproportional effect on the performance and almost no effect on the total size. 
    \item \textit{Erdős-Rényi:} As introduced in \cite{Mocanu2018}, $s^l$ scales with $1- \frac{n^{l-1}+n^{l}}{n^{l-1} * n^{l}}$, where $n^l$ denotes number of neurons at layer $l$. This enables the number of connections in a sparse layer to scale with the sum of the number of output and input channels.
    \item \textit{Erdős-Rényi-Kernel (ERK):} This method modifies the original Erdős-Rényi formulation by including the kernel dimensions in the scaling factors. In other words, the number of parameters of the sparse convolutional layers are scaled proportional to $1- \frac{n^{l-1}+n^{l}+w^l+h^l}{n^{l-1} * n^{l}*w^l*h^l}$, where $w^l$ and $h^l$ are the width and the height of the $l$'th convolutional kernel. Sparsity of the fully connected layers scale as in the original Erdős-Rényi formulation. Similar to Erdős-Rényi, ERK allocates higher sparsities to the layers with more parameters while allocating lower sparsities to the smaller ones.
\end{enumerate}

In all methods, the bias and batch-norm parameters are kept dense, since these parameters scale with total number of neurons and have a negligible effect on the total model size. 

\textbf{(2) Update Schedule.} The update schedule is defined by the following parameters: (1)$\Delta T$: the number of iterations between sparse connectivity updates, (2) $T_{end}$: the iteration at which to stop updating the sparse connectivity, (3) $\alpha$: the initial fraction of connections updated and (4) $f_{decay}$: a function, invoked every $\Delta T$ iterations until $T_{end}$, possibly decaying the fraction of updated connections over time. For the latter, as in \citet{dettmers2019}, we use \textit{cosine} annealing, as we find it slightly outperforms the other methods considered. 
$$f_{decay}(t;~\alpha,~T_{end})=\frac{\alpha}{2}\left(1+cos\left(\frac{t\pi}{T_{end}}\right)\right)$$

Results obtained with other annealing functions, such as \textit{constant} and \textit{inverse power}, are presented in Appendix \ref{app:schedules_other}.

\setlength{\tabcolsep}{0.3em}
\newcommand{\ci}[1]{\small{$\pm$#1}}

\textbf{(3) Drop criterion.} Every $\Delta T$ steps we drop the connections given by $ArgTopK(-|\theta^l|, f_{decay}(t;~\alpha,T_{end})(1-s^l) N^l)$, where $ArgTopK(v,~k)$ gives the indices of the top-$k$ elements of vector $v$.

\textbf{(4) Grow criterion.} The novelty of our method lies in how we grow new connections. We grow the connections with highest magnitude gradients, $ArgTopK_{i \notin \theta^l \setminus \mathbb{I}_{active}}(|\nabla_{\Theta^l} L_t|,~k)$, where  $\theta^l \setminus \mathbb{I}_{active}$ is the set of active connections remaining after step (3). Newly activated connections are \textbf{initialized to zero} and therefore don't affect the output of the network. However they are expected to receive gradients with high magnitudes in the next iteration and therefore reduce the loss fastest. We attempted using other initialization like random values or small values along the gradient direction for the activated connections, however these experiments didn't provide better results. 

This procedure can be applied to each layer in sequence and the dense gradients can be discarded immediately after selecting the top connections.  If a layer is too large to store the full gradient with respect to the weights, then the gradients can be calculated in an online manner and only the top-k gradient values are stored. As long as $\Delta T > \frac{1}{1-s}$, the extra work of calculating dense gradients is amortized and still proportional to $1-S$. This is in contrast to the method of~\cite{dettmers2019}, which requires calculating and storing the full gradients at each optimization step.
\begin{algorithm}[h]
\caption{RigL}
\label{alg:rigl}
\begin{algorithmic}
 \STATE {\bfseries Input:} Network $f_\Theta$, dataset $D$ \\
 \qquad\quad Sparsity Distribution: $\mathbb{S}=\{s^1, \dots, s^L\}$ \\
 \qquad\quad Update Schedule: $\Delta T$, $T_{end}$, $\alpha$, $f_{decay}$ 
 \STATE $\theta \gets$ Randomly sparsify $\Theta$ using $\mathbb{S}$ 
 \FOR{each training step $t$}
    \STATE Sample a batch $B_t \sim D$
    \STATE $L_t =\sum_{i \sim B_t} L((f_{\theta}(x_i),~y_i) $
    \IF {$t \ (\mathrm{mod}\ \Delta T)== 0$ and  $t < T_{end}$}
        \FOR{each layer $l$}
            \STATE $k = f_{decay}(t;~\alpha,T_{end})(1-s^l)N^l$ 
            \STATE $\mathbb{I}_{active} = ArgTopK(-|\theta^l|,~k)$ 
            \STATE $\mathbb{I}_{grow} = ArgTopK_{i \notin \theta^l \setminus \mathbb{I}_{active}}(|\nabla_{\Theta^l} L_t|,~k)$ 
            \STATE $\theta \gets$ Update connections $\theta$ using $\mathbb{I}_{active}$ and $\mathbb{I}_{grow}$
        \ENDFOR
    \ELSE
        \STATE $\theta = \theta - \alpha \nabla_{\theta}L_t$
    \ENDIF
 \ENDFOR
\end{algorithmic}
\end{algorithm}


\section{Empirical Evaluation}
\label{sec:experiments}

Our experiments include image classification using CNNs on the ImageNet-2012~\citep{imagenet} and CIFAR-10~\citep{cifar10} datasets and character based language modeling using RNNs with the WikiText-103 dataset \citep{wikitext103}. We repeat all of our experiments 3 times and report the mean and standard deviation. We use the TensorFlow Model Pruning library \citep{gupta2018} for our pruning baselines. A Tensorflow \citep{tensorflow2015} implementation of our method along with three other baselines (SET, SNFS, SNIP) and checkpoints of our models can be found at \href{https://github.com/google-research/rigl}{github.com/google-research/rigl}. You can also check the reproducibility report about our work at \href{https://github.com/varun19299/rigl-reproducibility}{https://github.com/varun19299/rigl-reproducibility} \citep{sundar2021reproducibility}.

For all dynamic sparse training methods (SET, SNFS, \textit{RigL}), we use the same update schedule with $\Delta T=100$ and $\alpha=0.3$ unless stated otherwise. Corresponding hyper-parameter sweeps can be found in Section \ref{sec:experiments_ablation}. We set the momentum value of SNFS to 0.9 and investigate other values in Appendix \ref{app:momentum}. We observed that stopping the mask updates prior to the end of training yields slightly better performance; therefore, we set $T_{end}$ to 25k for ImageNet-2012 and 75k for CIFAR-10 training which corresponds to roughly 3/4 of the full training. 

The default number of training steps used for training dense networks might not be optimal for sparse training with dynamic connectivity. In our experiments we observe that sparse training methods benefit significantly from increased training steps. When increasing the training steps by a factor $M$, the anchor epochs of the learning rate schedule and the end iteration of the mask update schedule are also scaled by the same factor; we indicate this scaling with a subscript (e.g. RigL$_{M\times}$). 

Additionally, in Appendix \ref{app:mnist}, we compare \textit{RigL} with structured pruning algorithms and in Appendix \ref{app:lottery} we show that solutions found by \textit{RigL} are not lottery tickets.
\subsection{ImageNet-2012 Dataset}
In all experiments in this section, we use SGD with momentum as our optimizer. We set the momentum coefficient of the optimizer to 0.9, $L_2$ regularization coefficient to 0.0001, and label smoothing \citep{labelsmooth} to 0.1. The learning rate schedule starts with a linear warm up reaching its maximum value of 1.6 at epoch 5 which is then dropped by a factor of 10 at epochs 30, 70 and 90. We train our networks with a batch size of 4096 for 32000 steps which roughly corresponds to 100 epochs of training. Our training pipeline uses standard data augmentation, which includes random flips and crops. 
\begin{figure*}[t]
\centering
\begin{minipage}{.63\textwidth}
  \centering
  \begin{tabular}{c|p{4em}|p{3em}|p{3em}||p{4em}|p{3em}|p{3em}}
    \toprule
     Method & Top-1 \newline Accuracy& FLOPs \newline (Train) & FLOPs\newline (Test) & Top-1 \newline Accuracy& FLOPs \newline (Train) & FLOPs\newline (Test) \\\toprule
    Dense & 76.8\ci{0.09} & 1x \small{(3.2e18)} & 1x \small{(8.2e9)} \\\hline
        & \multicolumn{3}{c ||}{S=0.8} & \multicolumn{3}{ c }{S=0.9}\\\hline \hline
    Static & 70.6\ci{0.06} & 0.23x & 0.23x & 65.8\ci{0.04} & 0.10x & 0.10x \\
    SNIP & 72.0\ci{0.10} & 0.23x & 0.23x & 67.2\ci{0.12} & 0.10x & 0.10x \\
    Small-Dense & 72.1\ci{0.12} & 0.20x & 0.20x & 68.9\ci{0.10} & 0.12x & 0.12x \\
    SET & 72.9\ci{0.39} & 0.23x & 0.23x & 69.6\ci{0.23} & 0.10x & 0.10x \\
    RigL & 74.6\ci{0.06} & 0.23x & 0.23x & 72.0\ci{0.05} & 0.10x & 0.10x \\
    Small-Dense$_{5\times}$ & 73.9\ci{0.07} & 1.01x & 0.20x & 71.3\ci{0.10} & 0.60x & 0.12x \\
    RigL$_{5\times}$ & \textbf{76.6\ci{0.06}} & 1.14x & 0.23x & \textbf{75.7\ci{0.06}} & 0.52x & 0.10x \\\hline
    Static (ERK) & 72.1\ci{0.04} & 0.42x & 0.42x & 67.7\ci{0.12} & 0.24x & 0.24x \\
    DSR* & 73.3 & 0.40x & 0.40x & 71.6 & 0.30x & 0.30x \\
    RigL (ERK) & 75.1\ci{0.05} & 0.42x & 0.42x & 73.0\ci{0.04} & 0.25x & 0.24x \\
    RigL$_{5\times}$ (ERK) & \textbf{77.1\ci{0.06}} & 2.09x & 0.42x & \textbf{76.4\ci{0.05}} & 1.23x & 0.24x \\\hline \hline
    SNFS* & 74.2 & n/a & n/a & 72.3 & n/a & n/a \\
    SNFS (ERK) & 75.2\ci{0.11} & 0.61x & 0.42x & 72.9\ci{0.06} & 0.50x & 0.24x \\
    Pruning* & 75.6 & 0.56x & 0.23x & 73.9 & 0.51x & 0.10x \\
    Pruning$_{1.5\times}$* & \textbf{76.5} & 0.84x & 0.23x & \textbf{75.2} & 0.76x & 0.10x \\
    DNW* & 76 & n/a & n/a & 74 & n/a & n/a \\
    \bottomrule
    \end{tabular}
\end{minipage}
\begin{minipage}{.35\textwidth}
  \centering
  \includegraphics[width=.98\linewidth]{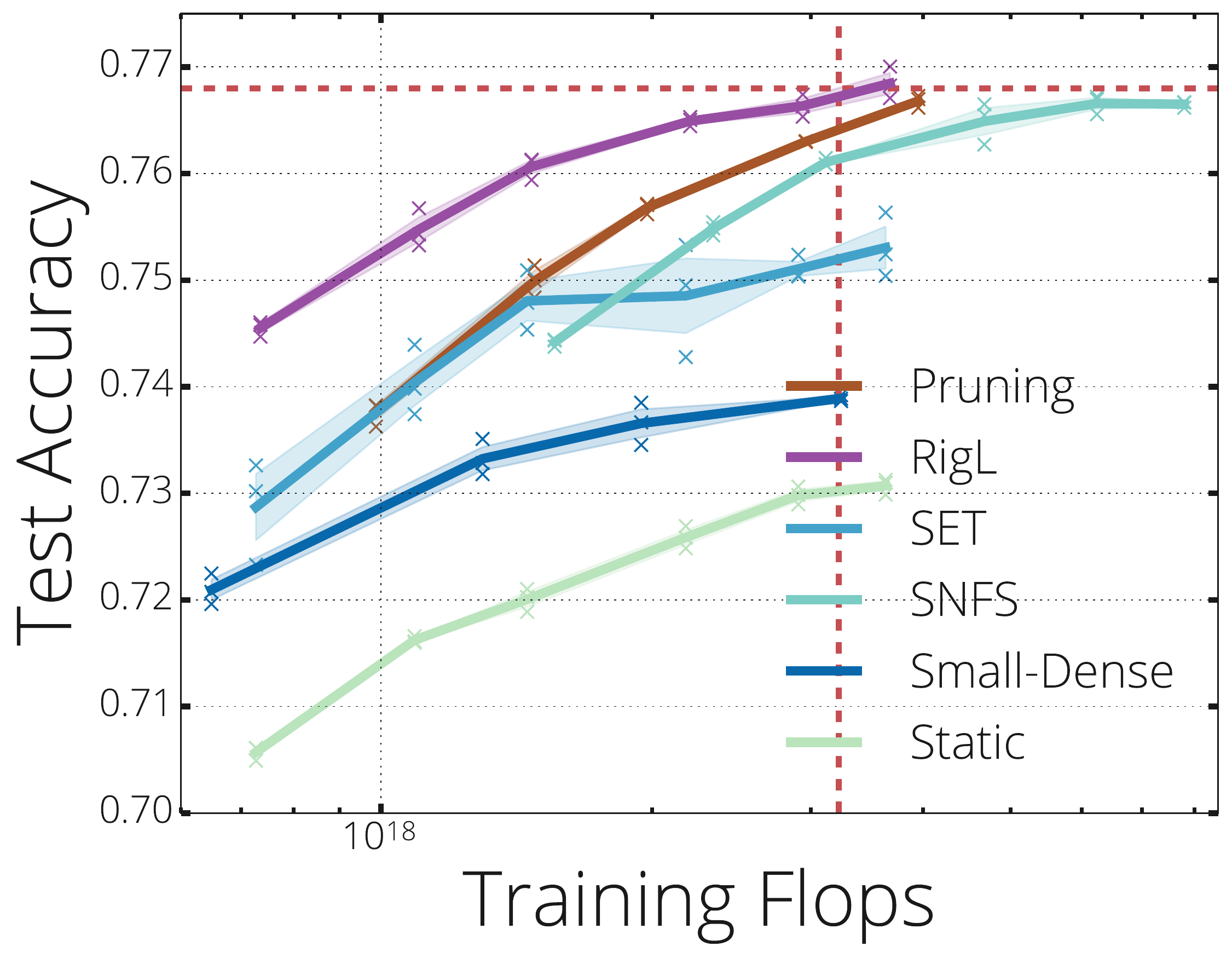}
  \includegraphics[width=.98\linewidth]{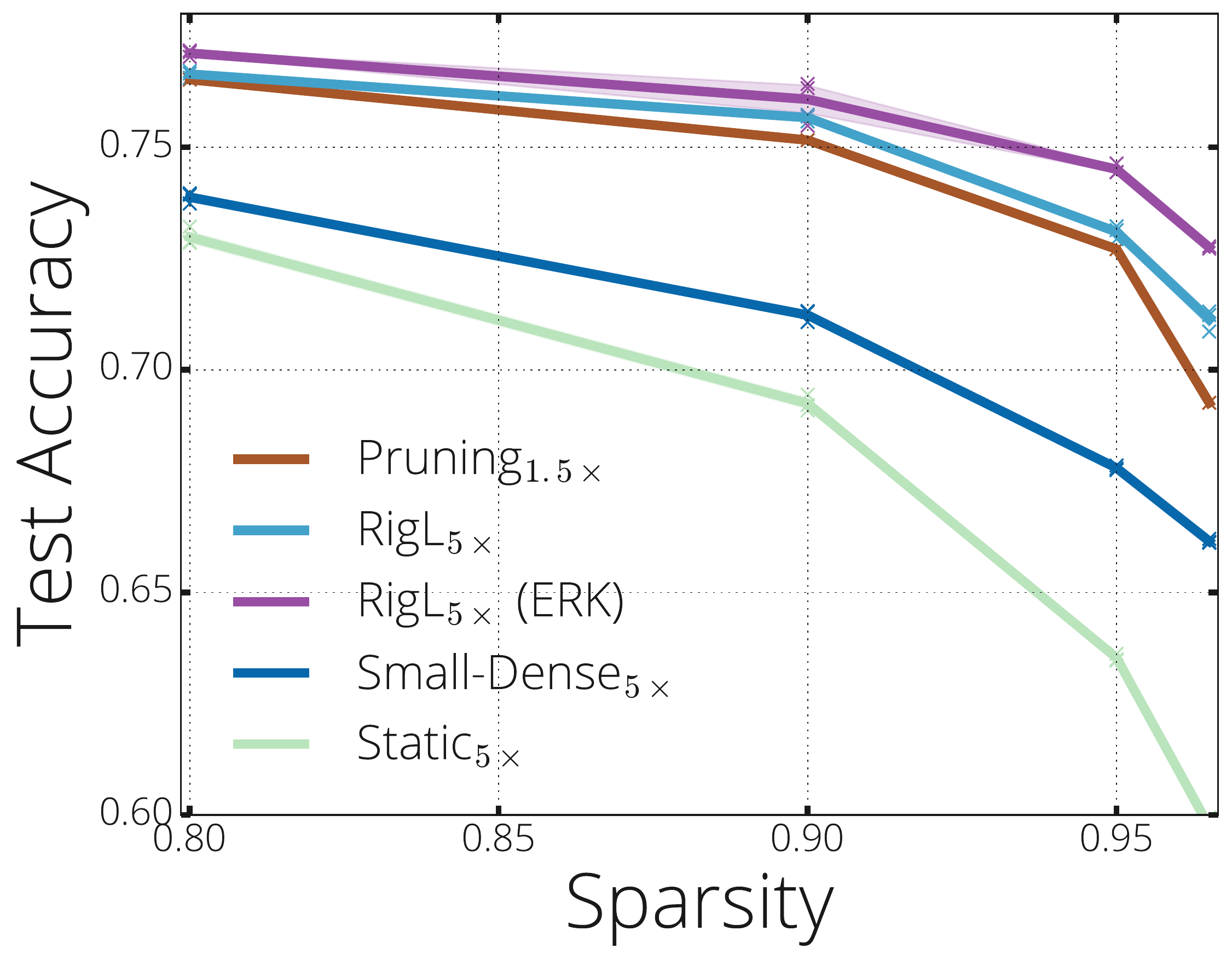}
\end{minipage}%
\caption{\textbf{(left)} Performance and cost of training 80\% and 90\% sparse ResNet-50s on the Imagenet-2012 classification task. We report FLOPs needed for training and test (inference on single sample) and normalize them with the FLOPs of a dense model. To make a fair comparison we assume pruning algorithm utilizes sparsity during the training (see Appendix \ref{app:flops} for details on how FLOPs are calculated). Methods with superscript `*' indicates reported results in corresponding papers (except DNW results, which is obtained from \citep{kusupati2020str}). Pruning results are obtained from \cite{gale2019state}. \textbf{(top-right)} Performance of sparse training methods on training 80\% sparse ResNet-50 with uniform sparsity distribution. Points at each curve correspond to the individual training runs with training multipliers from 1 to 5 (except pruning which is scaled between 0.5 and 2). The number of FLOPs required to train a standard dense ResNet-50 along with its performance is indicated with a dashed red line. ~\textbf{(bottom-right)} Performance of \textit{RigL} at different sparsity levels with extended training.}
\label{fig:resnet}
\end{figure*}

\subsubsection{ResNet-50}
\label{sec:experiments_ResNet}
Figure \ref{fig:resnet}-top-right summarizes the performance of various methods on training an 80\% sparse ResNet-50. We also train small dense networks with equivalent parameter count. All sparse networks use a uniform layer-wise sparsity distribution unless otherwise specified and a cosine update schedule ($\alpha=0.3$, $\Delta T=100$). Overall, we observe that the performance of all methods improves with training time; thus, for each method we run extended training with up to 5$\times$ the training steps of the original.


As noted by \citet{gale2019state}, \citet{Evci2019}, \citet{frankle2019}, and \citet{Mostafa2019}, training a network with fixed sparsity from scratch (\textit{Static}) leads to inferior performance. Training a dense network with the same number of parameters (\textit{Small-Dense}) gets better results than \textit{Static}, but fails to match the performance of dynamic sparse models. SET improves the performance over \textit{Small-Dense}, however saturates around 75\% accuracy indicating the limits of growing new connections randomly. Methods that use gradient information to grow new connections (\textit{RigL} and SNFS) obtain higher accuracies, but {\em RigL} achieves the highest accuracy and does so while consistently requiring fewer FLOPs than the other methods.

\begin{figure*}
\centering
\begin{minipage}{.44\textwidth}
  \centering
  \includegraphics[width=.95\linewidth]{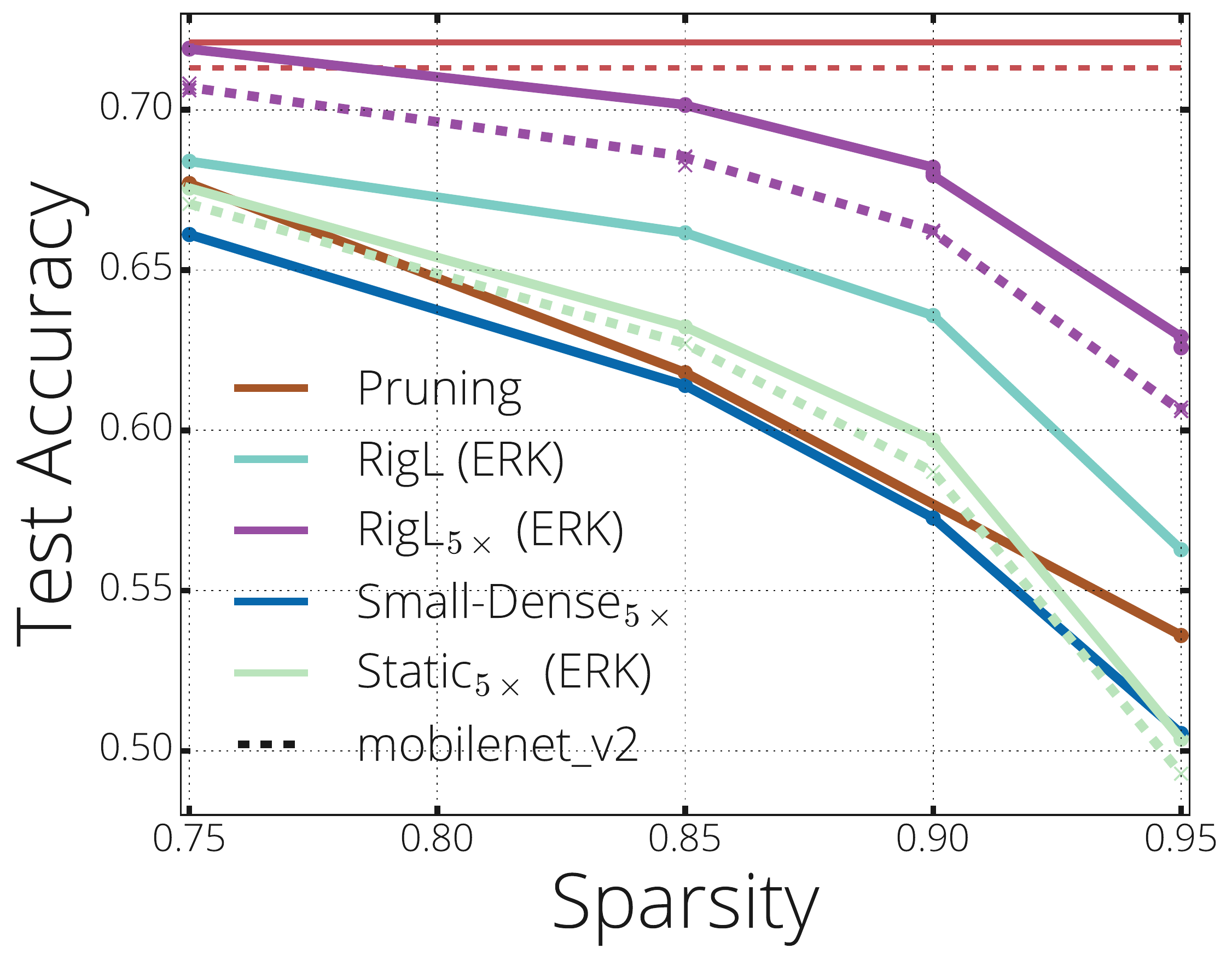}
\end{minipage}%
\begin{minipage}{.56\textwidth}
 \centering
\begin{tabular}{r|l|c|c}
 
S &Method & Top-1  & FLOPs \\
\hline \multirow{2}{1.8em}{\small0.75} & Small-Dense$_{5\times}$ & 66.0\ci{0.11} & 0.23x \\ 
 & Pruning (Zhu) & 67.7 & 0.27x \\
 & RigL$_{5\times}$ & 71.5\ci{0.06} & 0.27x \\
 & RigL$_{5\times}$ (ERK) & \textbf{71.9\ci{0.01}} & 0.52x \\
\hline \multirow{2}{1.8em}{\small0.9} & Small-Dense$_{5\times}$ & 57.7\ci{0.34} & 0.09x \\ 
 & Pruning (Zhu) & 61.8 & 0.12x \\
 & RigL$_{5\times}$ & 67.0\ci{0.17} & 0.12x \\
 & RigL$_{5\times}$ (ERK) & \textbf{68.1\ci{0.11}} & 0.27x \\
\hline\hline & Dense & 72.1\ci{0.17} & 1x (1.1e9) \\
\hline \multirow{2}{1.8em}{\small0.75} & Big-Sparse$_{5\times}$ & 76.4\ci{0.05} & 0.98x \\ 
 & Big-Sparse$_{5\times}$ (ERK) &\textbf{ 77.0\ci{0.08}} & 1.91x \\
\end{tabular}
 
\end{minipage}
\caption{\textbf{(left)} {\em RigL} significantly improves the performance of sparse MobileNets (v1 and v2) on ImageNet-2012 dataset and exceeds the \textit{pruning} results reported by \cite{gupta2018}. Performance of the dense MobileNets are indicated with red lines. \textbf{(right)} Performance of sparse MobileNet-v1 architectures presented with their inference FLOPs. Networks with \textit{ERK} distribution get better performance with the same number of parameters but take more FLOPs to run. Training wider sparse models with {\em RigL} (\textit{Big-Sparse}) yields a significant performance improvement over the dense model.}
\label{fig:mobile}
\vskip -0.1in
\end{figure*}
Given that different applications or scenarios might require a limit on the number of FLOPs for inference, we investigate the performance of our method at various sparsity levels. As mentioned previously, one strength of our method is that its resource requirements are constant throughout training and we can choose the level of sparsity that fits our training and/or inference constraints. In Figure \ref{fig:resnet}-bottom-right we show the performance of our method at different sparsities and compare them with the pruning results of \cite{gale2019state}, which uses 1.5x training steps, relative to the original 32k iterations. To make a fair comparison with regards to FLOPs, we scale the learning schedule of all other methods by 5x. Note that even after extending the training, it takes less FLOPs to train sparse networks using {\em RigL} compared to the pruning method\footnote{Except for the 80\% sparse {\em RigL}-ERK}.

\textit{RigL}, our method with constant sparsity distribution, \textbf{exceeds} the performance of magnitude based iterative pruning in all sparsity levels while requiring less FLOPs to train. Sparse networks that use \textit{Erdős-Renyi-Kernel (ERK)} sparsity distribution obtains even greater performance. For example ResNet-50 with 96.5\% sparsity achieves a remarkable 72.75\% Top-1 Accuracy, around 3.5\% higher than the extended magnitude pruning results reported by \cite{gale2019state}. As observed earlier, smaller dense models (with the same number of parameters) or sparse models with a static connectivity can not perform at a comparable level.

A more fine grained comparison of sparse training methods is presented in Figure \ref{fig:resnet}-left. Methods using uniform sparsity distribution and whose FLOP/memory footprint scales directly with (1-S) are placed in the first sub-group of the table. The second sub-group includes DSR and networks with ERK sparsity distribution which require a higher number of FLOPs for inference with same parameter count. The final sub-group includes methods that require the space and the work proportional to training a dense model.

\subsubsection{MobileNet}
\label{sec:experiments_mnet}
MobileNet is a compact architecture that performs remarkably well in resource constrained settings. Due to its compact nature with separable convolutions it is known to be difficult to sparsify without compromising performance \citep{gupta2018}. In this section we apply our method to MobileNet-v1 \citep{mobilenetv1} and MobileNet-v2 \citep{mobilenetv2}. Due to its low parameter count we keep the first layer and depth-wise convolutions dense. We use ERK or Uniform sparsity distributions to sparsify the remaining layers. We calculate sparsity fractions in this section over pruned layers and real sparsities (when first layer and depth-wise convolutions are included) are slightly lower than the reported values (i.e. 74.2, 84.1, 89, 94 for 75, 85, 90, 95 \% sparsity).

The performance of sparse MobileNets trained with {\em RigL} as well as the baselines are shown in Figure \ref{fig:mobile}. We extend the training (5x of the original number of steps) for all runs in this section. \textit{RigL} trains 75\% sparse MobileNets with no loss in performance. Performance starts dropping after this point, though {\em RigL} consistently gets the best results by a large margin.

Figure \ref{fig:resnet}-top-right and Figure \ref{fig:mobile}-left show that the sparse models are more accurate than the dense models with the same number of parameters, corroborating the results of \citet{kalchbrenner2018}. To validate this point further, we train a sparse MobileNet-v1 with width multiplier of 1.98 and constant sparsity of 75\%, which has the same FLOPs and parameter count as the dense baseline. Training this network with {\em RigL} yields an impressive \textbf{4.3\% absolute improvement} in Top-1 Accuracy demonstrating the exciting potential of sparse networks at increasing the performance of widely-used dense models.
\begin{figure*}[!t]
\centering
\begin{minipage}{.5\textwidth}
  \centering
  \includegraphics[width=.95\linewidth]{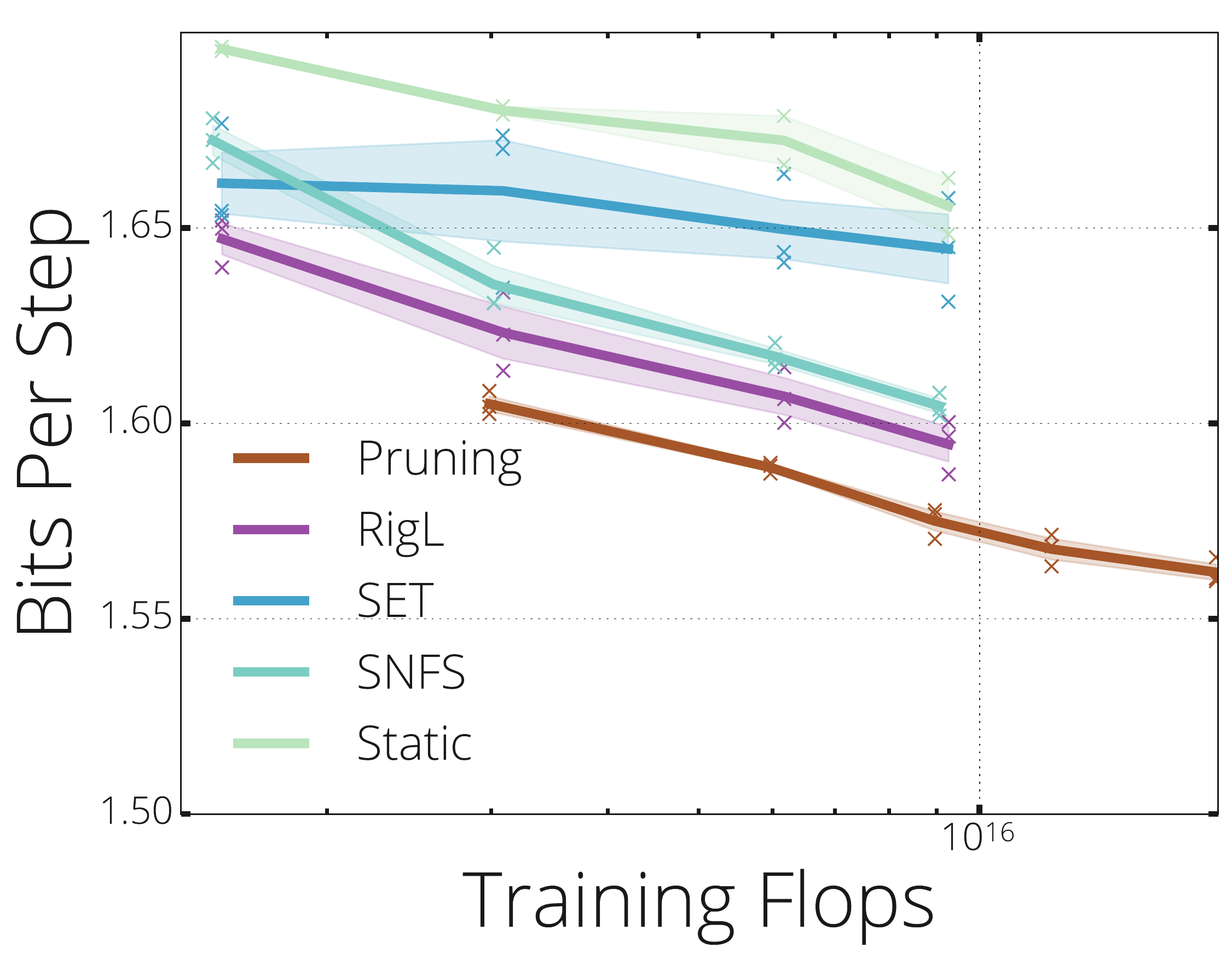}
\end{minipage}%
\begin{minipage}{.5\textwidth}
  \centering
  \includegraphics[width=.9\linewidth]{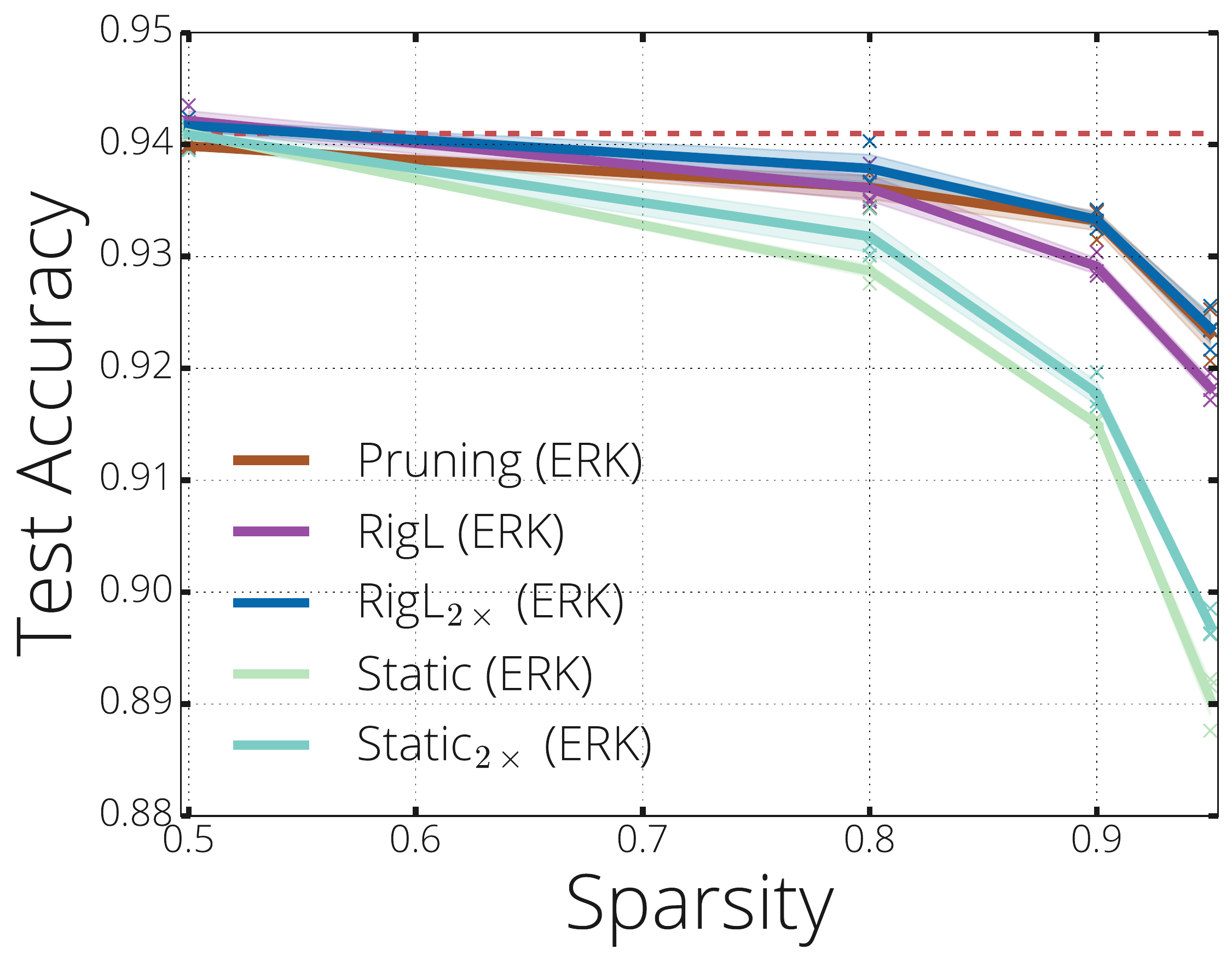}
\end{minipage}
\caption{\textbf{(left)} Final validation loss of various sparse training methods on character level language modeling task. Cross entropy loss is converted to bits (from nats). \textbf{(right)} Test accuracies of sparse WideResNet-22-2's on CIFAR-10 task.}
\label{fig:wikichar_cifar10}
\end{figure*}

\subsection{Character Level Language Modeling}
\label{sec:chargru}

Most prior work has only examined sparse training on vision networks \footnote{The exception being the work of \citet{Bellec2017}}.  To fully understand these techniques it is important to examine different architectures on different datasets.  \citet{kalchbrenner2018} found sparse GRUs~\citep{GRU} to be very effective at modeling speech, however the dataset they used is not available.  We choose a proxy task with similar characteristics (dataset size and vocabulary size are approximately the same) - character level language modeling on the publicly available WikiText-103~\citep{wikitext103} dataset.

\begin{figure*}[b]
\centering
\begin{minipage}{.5\textwidth}
  \centering
  \includegraphics[width=.95\linewidth]{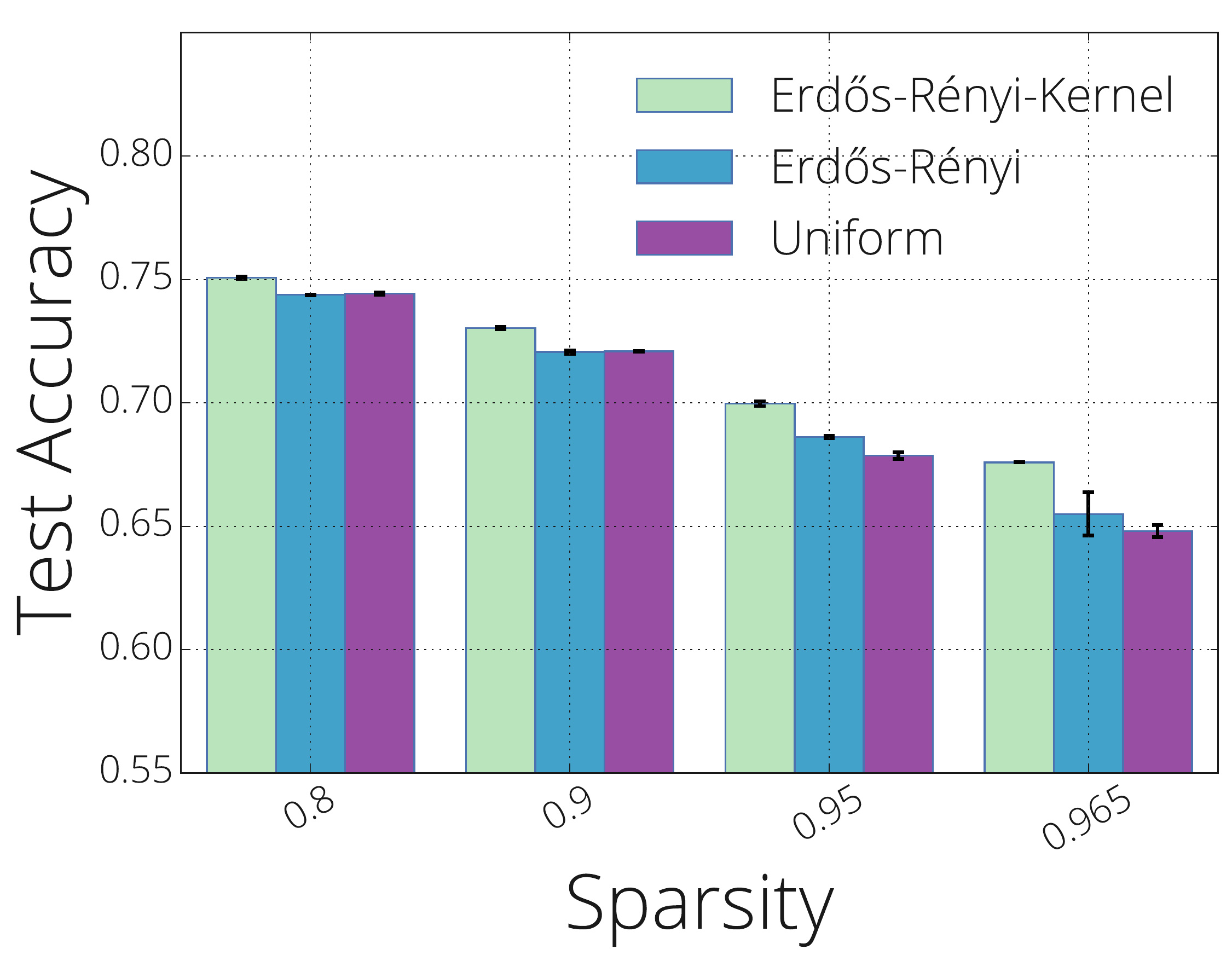}
\end{minipage}%
\begin{minipage}{.5\textwidth}
  \centering
  \includegraphics[width=.9\linewidth]{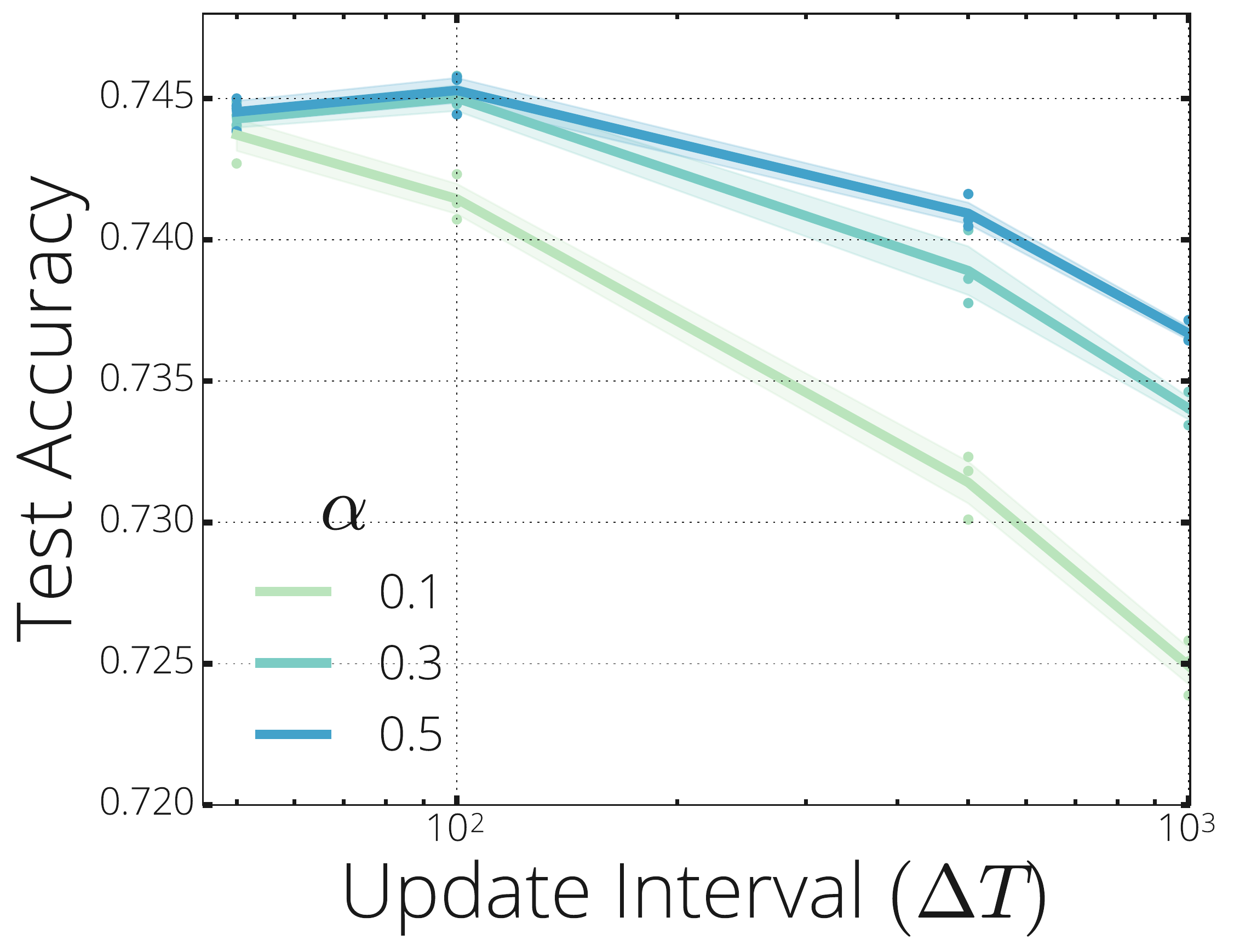}
\end{minipage}
\vskip -0.05in
\caption{Effect of \textbf{(left)} sparsity distribution and \textbf{(right)} update schedule ($\Delta T$, $\alpha$) on the final performance.  }
\label{fig:ablation}
\end{figure*}

Our network consists of a shared embedding with dimensionality 128, a vocabulary size of 256, a GRU with a state size of 512, a readout from the GRU state consisting of two linear layers with width 256 and 128 respectively.  We train the next step prediction task with the cross entropy loss using the Adam optimizer. The remaining hyper-parameters are reported in Appendix \ref{app:chargru}.

In Figure \ref{fig:wikichar_cifar10}-left we report the validation bits per step of various solutions at the end of the training. For each method we perform extended runs to see how the performance of each method scales with increased training time. As observed before, SET performs worse than the other dynamic training methods and its performance improves only slightly with increased training time. On the other hand the performance of {\em RigL} and SNFS continuously improves with more training steps. Even though {\em RigL} exceeds the performance of the other sparse training approaches it fails to match the performance of pruning in this setting, highlighting an important direction for future work. 

\subsection{WideResNet-22-2 on CIFAR-10}
\label{sec:cifar10}
We also evaluate the performance of {\em RigL} on the CIFAR-10 image classification benchmark. We train a Wide Residual Network \citep{wideresnet} with 22 layers using a width multiplier of 2 for 250 epochs (97656 steps). The learning rate starts at 0.1 which is scaled down by a factor of 5 every 30,000 iterations. We use an L2 regularization coefficient of 5e-4, a batch size of 128 and a momentum coefficient of 0.9. We use the default mask update interval for {\em RigL} ($\Delta T=100$) and the default ERK sparsity distribution. Results with other mask update intervals and sparsity distributions yield similar results. These can be found in Appendix \ref{app:cifar10}.

The final accuracy of {\em RigL} for various sparsity levels is presented in Figure \ref{fig:wikichar_cifar10}-right. The dense baseline obtains 94.1\% test accuracy; surprisingly, some of the 50\% sparse networks generalize better than the dense baseline demonstrating the regularization aspect of sparsity. With increased sparsity, we see a performance gap between the \textit{Static} and \textit{Pruning} solutions. Training static networks longer seems to have limited effect on the final performance. On the other hand, {\em RigL} matches the performance of pruning with only a fraction of the resources needed for training. 
\subsection{Analyzing the performance of {\em RigL}}
\label{sec:experiments_ablation}

In this section we study the effect of \textit{sparsity distributions} and \textit{update schedules} on the performance of our method. The results for SET and SNFS are similar and are discussed in Appendices \ref{app:initmethods} and \ref{app:schedules}. Additionally, we investigate the energy landscape of sparse ResNet-50s and show that dynamic connectivity provided by {\em RigL} helps escaping sub-optimal solutions found by static training. 

\begin{figure*}
\centering
\begin{minipage}{.4\textwidth}
  \centering
  \includegraphics[width=.95\linewidth]{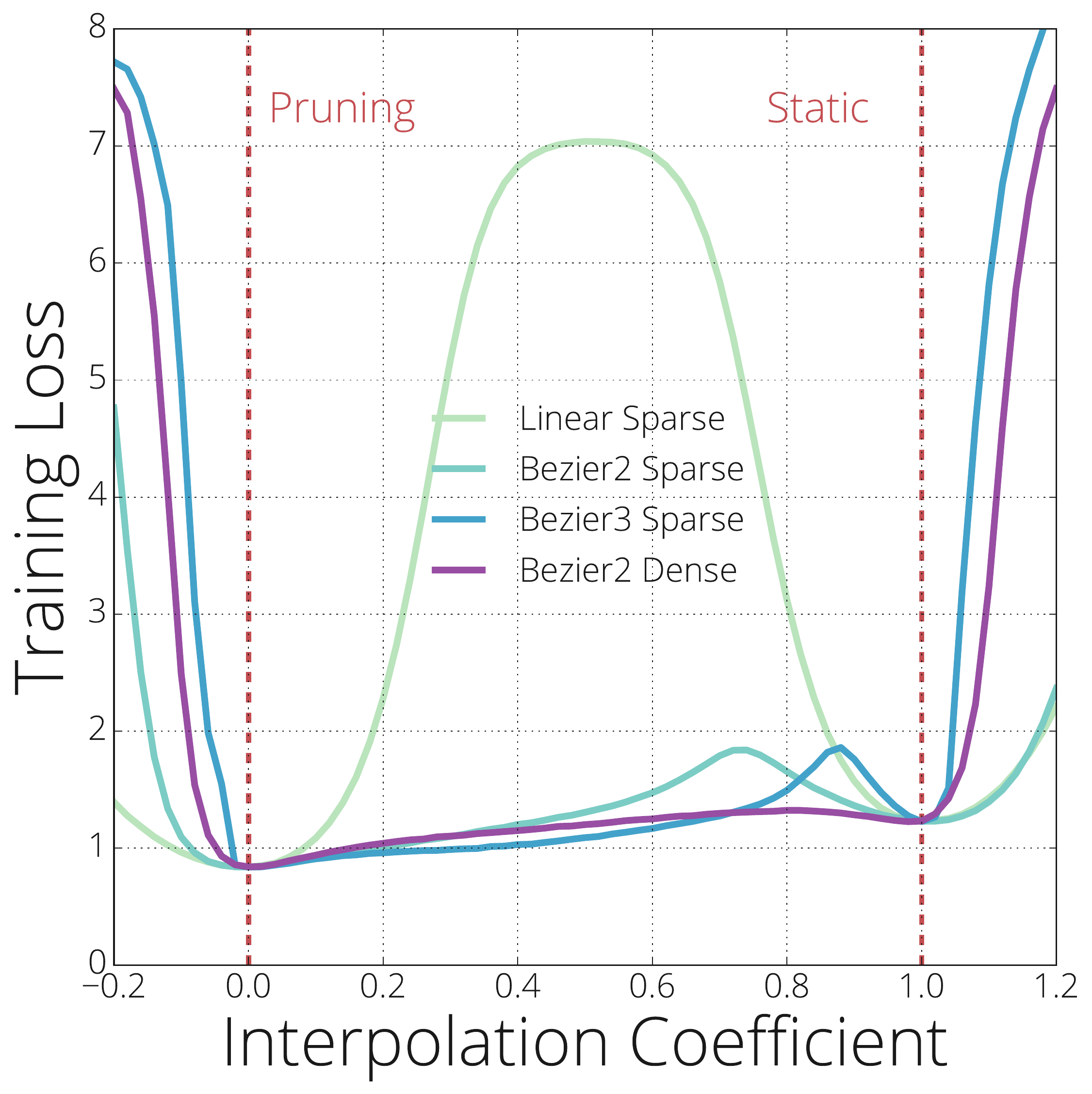}
\end{minipage}%
\begin{minipage}{.6\textwidth}
  \centering
  \includegraphics[width=.95\linewidth]{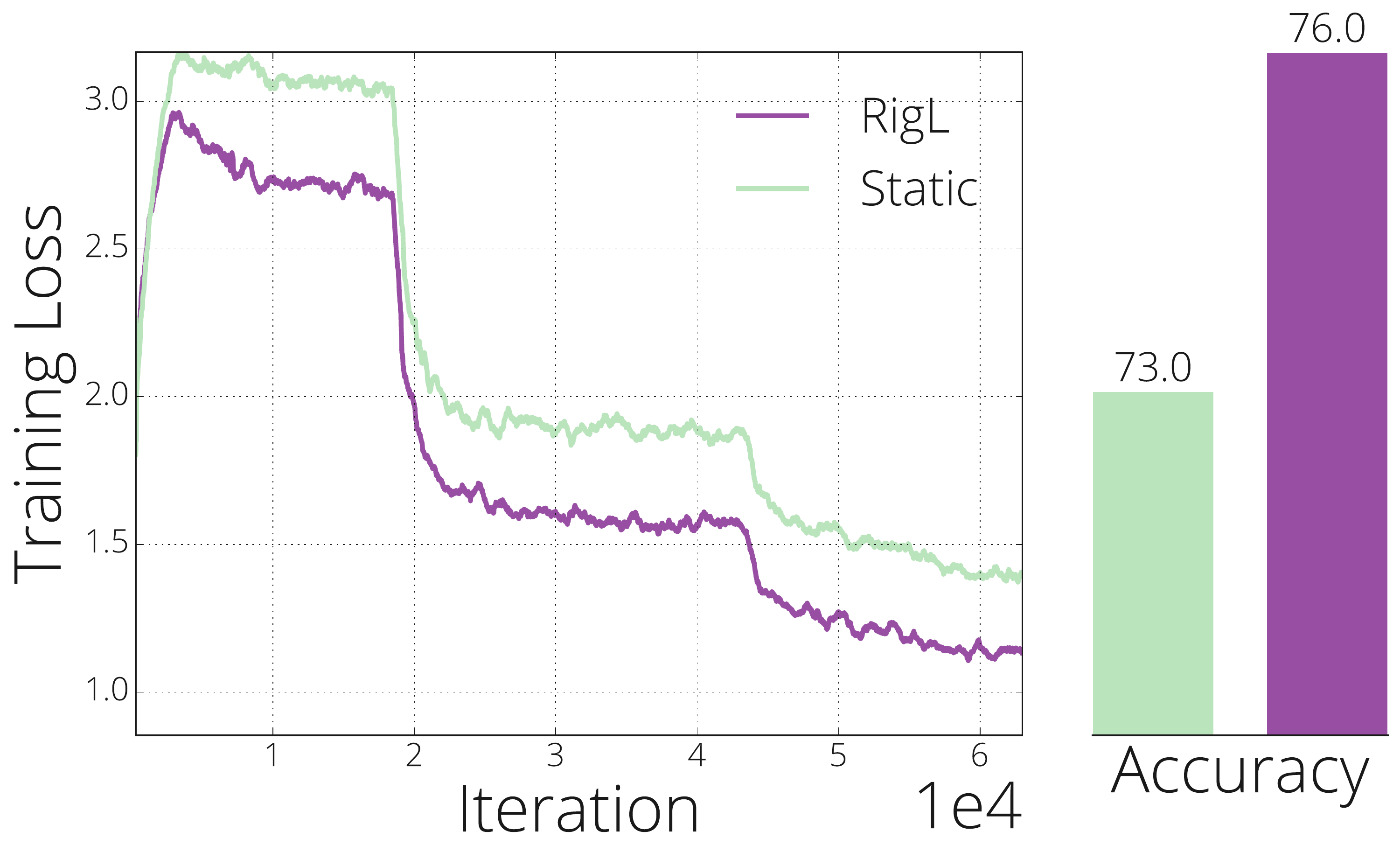}
\end{minipage}
\caption{\textbf{(left)} Training loss evaluated at various points on interpolation curves between a magnitude pruning model (0.0) and a model trained with static sparsity (1.0). \textbf{(right)} Training loss of {\em RigL} and {\em Static} methods starting from the static sparse solution, and their final accuracies.}
\label{fig:discussion}
\end{figure*}

\textbf{Effect of Sparsity Distribution:}
Figure \ref{fig:ablation}-left shows how the sparsity distribution affects the final test accuracy of sparse ResNet-50s trained with {\em RigL}. Erdős-Rényi-Kernel (ERK) performs consistently better than the other two distributions. ERK automatically allocates more parameters to the layers with few parameters by decreasing their sparsities\footnote{see Appendix \ref{app:sparsities} for exact layer-wise sparsities.}. This reallocation seems to be crucial for preserving the capacity of the network at high sparsity levels where ERK outperforms other distributions by a greater margin. Though it performs better, the ERK distribution requires approximately twice as many FLOPs compared to a uniform distribution. This highlights an interesting trade-off between accuracy and computational efficiency where better performance is obtained by increasing the number of FLOPs required to evaluate the model. This trade-off also highlights the importance of reporting non-uniform sparsities along with respective FLOPs when two networks of same sparsity (parameter count)  are compared.

\textbf{Effect of Update Schedule and Frequency:}
In Figure \ref{fig:ablation}-right, we evaluate the performance of our method on update intervals $\Delta T\in [50,100,500,1000]$ and initial drop fractions $\alpha \in [0.1,0.3,0.5]$. The best accuracies are obtained when the mask is updated every 100 iterations with an initial drop fraction of 0.3 or 0.5. Notably, even with infrequent update intervals (e.g. every 1000 iterations), \textit{RigL} performs above 73.5\%.



\textbf{Effect of Dynamic connections:}
\citet{frankle2019} and \citet{Mostafa2019} observed that static sparse training converges to a solution with a higher loss than dynamic sparse training. In Figure~\ref{fig:discussion}-left we examine the loss landscape lying between a solution found via static sparse training and a solution found via pruning to understand whether the former lies in a basin isolated from the latter. Performing a linear interpolation between the two reveals the expected result -- a high-loss barrier -- demonstrating that the loss landscape is not trivially connected. However, this is only one of infinitely many paths between the two points \citep{Garipov2018, Draxler2018} and does not imply the nonexistence of such a path. For example \citet{Garipov2018} showed different dense solutions lie in the same basin by finding $2^{nd}$ order B\'ezier curves with low energy between the two solutions. Following their method, we attempt to find quadratic and cubic B\'ezier curves between the two sparse solutions. Surprisingly, even with a cubic curve, we fail to find a path without a high-loss barrier. These results suggest that static sparse training can get stuck at local minima that are isolated from better solutions. On the other hand, when we optimize the quadratic B\'ezier curve across the full dense space we find a near-monotonic path to the improved solution, suggesting that allowing new connections to grow yields greater flexibility in navigating the loss landscape.

In Figure~\ref{fig:discussion}-right we train {\em RigL} starting from the sub-optimal solution found by static sparse training, demonstrating that it is able to escape the local minimum, whereas re-training with static sparse training cannot. {\em RigL} first removes connections with the smallest magnitudes since removing these connections have been shown to have a minimal effect on the loss \citep{han2015learning, evci2018}. Next, it activates connections with the high gradients, since these connections are expected to decrease the loss fastest. In Appendix \ref{app:losssurface} we discuss the effect of \textit{RigL} updates on the energy landscape.

\section{Discussion \& Conclusion}
In this work we introduced \textit{RigL}, an algorithm for training sparse neural networks efficiently. For a given computational budget \textit{RigL} achieves higher accuracies than existing dense-to-sparse and sparse-to-sparse training algorithms.
\textit{RigL} is useful in three different scenarios: (1) To improve the accuracy of sparse models intended for deployment; (2) To improve the accuracy of large sparse models which can only be trained for a limited number of iterations; (3) Combined with sparse primitives to enable training of extremely large sparse models which otherwise would not be possible.

The third scenario is unexplored due to the lack of hardware and software support for sparsity.
Nonetheless, work continues to improve the performance of sparse networks on current hardware~\citep{AdaptiveSparseTilingGPU, Merrill_MergePath_SpMV}, and new types of hardware accelerators will have better support for parameter sparsity~\citep{snnramsparsehardware, myrtleunstructured, Liu2018MemoryEfficientDL, Han2016, eyeriss_sparse_accelerator}. \textit{RigL} provides the tools to take advantage of, and motivation for, such advances.

\subsubsection*{Acknowledgments}
We would like to thank Eleni Triantafillou, Hugo Larochelle, Bart van Merriënboer, Fabian Pedregosa, Joan Puigcerver, Nicolas Le Roux, Karen Simonyan for giving feedback on the preprint of the paper; Namhoon Lee for helping us verifying/debugging our SNIP implementation; Chris Jones for helping discovering/solving the distributed training bug; Dyllan McCreary for helping us improve the notation; D.C. Mocanu for helping us clarify related work timeline; Varun Sundar and Rajat Vadiraj Dwaraknath for choosing our work for the ML Reproducibility Challenge 2020.

\bibliography{example_paper}
\bibliographystyle{icml2020}
\clearpage
\appendix

\section{Effect of Mask Updates on the Energy Landscape}
\label{app:losssurface}
To update the connectivity of our sparse network, we first need to drop a fraction $d$ of the existing connections for each layer independently to create a budget for growing new connections. Like many prior works~\citep{Thimm95evaluatingpruning, sparse-connection-1997, exploring-sparsity-rnn, han2015learning}, we drop parameters with the smallest magnitude. The effectiveness of this simple criteria can be explained through the first order Taylor approximation of the loss $L$ around the current set of parameters $\theta$.

$$ \Delta L = L(\theta+\Delta \theta) - L(\theta) = \nabla_\theta L(\theta) \Delta \theta + R(||\Delta \theta||^2_2) $$

The main goal of dropping connections is to remove parameters with minimal impact on the output of the neural network and therefore on its loss. Since removing the connection $\theta_i$ corresponds to setting it to zero, it incurs a change of $\Delta\theta=-\theta_i$ in that direction and a change of $\Delta L_i = -\nabla_{\theta_i} L(\theta)\theta_i  + R(\theta_i^2)$ in the loss, where the first term is usually defined as the {\em saliency} of a connection. Saliency has been used as a criterion to remove connections~\citep{pruning-convnet-nvidia}, however it has been shown to produce inferior results compared to magnitude based removal, especially when used to remove multiple connections at once~\citep{evci2018}. In contrast, picking the lowest magnitude connections ensures a small remainder term in addition to a low saliency, limiting the damage we make when we drop connections. Additionally, we note that connections with small magnitude can only remain small if the gradient they receive during training is small, meaning that the saliency is likely small when the parameter itself is small.

After the removal of insignificant connections, we enable new connections that have the highest expected gradients. Since we initialize these new connections to zero, they are guaranteed to have high gradients in the proceeding iteration and therefore to reduce the loss quickly.  Combining this observation with the fact that {\em RigL} is likely to remove low gradient directions, ) and the results in Section \ref{sec:experiments_ablation}, suggests that {\em RigL} improves the energy landscape of the optimization by replacing flat dimensions with ones with higher gradient. This helps the optimization procedure escape saddle points.

\section{Comparison with Bayesian Structured Pruning Algorithms}
\label{app:mnist}
Structured pruning algorithms aim to remove entire neurons (or channels) instead of individual connections either at the end of, or throughout training. The final pruned network is a smaller dense network. \citet{Liu2018} demonstrated that these smaller networks could themselves be successfully be trained from scratch. This recasts structured pruning approaches as a limited kind of architecture search, where the search space is the size of each hidden layer. 

In this section we compare \textit{RigL} with three different structured pruning algorithms: SBP \citep{sbp2017}, L0 \citep{Louizos2018}, and VIB \citep{vib2018}. We show that starting from a random sparse network, \textit{RigL} finds compact networks with fewer parameters, that require fewer FLOPs for inference \emph{and} require fewer resources for training. This serves as a general demonstration of the effectiveness of unstructured sparsity.

\begin{table*}
    \centering
    \begin{tabular}{l|cccccc}\toprule
 
    Method      & Final  & Sparsity  & Training Cost & Inference Cost& Size & Error \\
          & Architecture  &   & (GFLOPs) & (KFLOPs) & (bytes) &  \\
\midrule
SBP	& 245-160-55	& 0.000	& 13521.6 (2554.8)	& 97.1	& 195100	& 1.6	 \\
L0	& 266-88-33	& 0.000	& 13521.6 (1356.4) & 53.3	& 107092	& 1.6	 \\
VIB	& 97-71-33	& 0.000	& 13521.6 (523) & 19.1	& 38696	& 1.6	 \\
RigL	& 408-100-69	& 0.870	& 482.0	& 12.6	& 31914	& 1.44 (1.48)	 \\
RigL+	& 375-62-51	& 0.886	& \textbf{206.3}	& \textbf{6.2}	& \textbf{16113}	& 1.57 (1.69)	 \\
\bottomrule    \end{tabular}
    \caption{Performance of various structured pruning algorithms on compressing three layer MLP on MNIST task. Cost of training the final architectures found by SBP, L0 and VIB are reported in parenthesis. \textit{RigL} finds more compact networks compared to structured pruning approaches.}
    \label{tab:app:mnist}
\end{table*}

For our setting we pick the standard LeNet 300-100 network with ReLU non-linearities trained on MNIST. In Table \ref{tab:app:mnist} we compare methods based on how many FLOPs they require for training and also how efficient the final architecture is. Unfortunately, none of the papers have released the code for reproducing the MLP results, so we therefore use the reported accuracies and calculate lower bounds for the the FLOPs used during training. For each method we assume that one training step takes as much as the dense 300-100 architecture and omit any additional operations each method introduces. We also consider training the pruned networks from scratch and report the training FLOPs required in parenthesis. In this setting, training FLOPs are significantly lower since the starting networks are have been significantly reduced in size.  We assume that following~\citep{Liu2018} the final networks can be trained from scratch, but we cannot verify this for these MLP networks since it would require knowledge of which pixels were dropped from the input. 
\begin{figure*}
\centering
  \includegraphics[width=.99\linewidth]{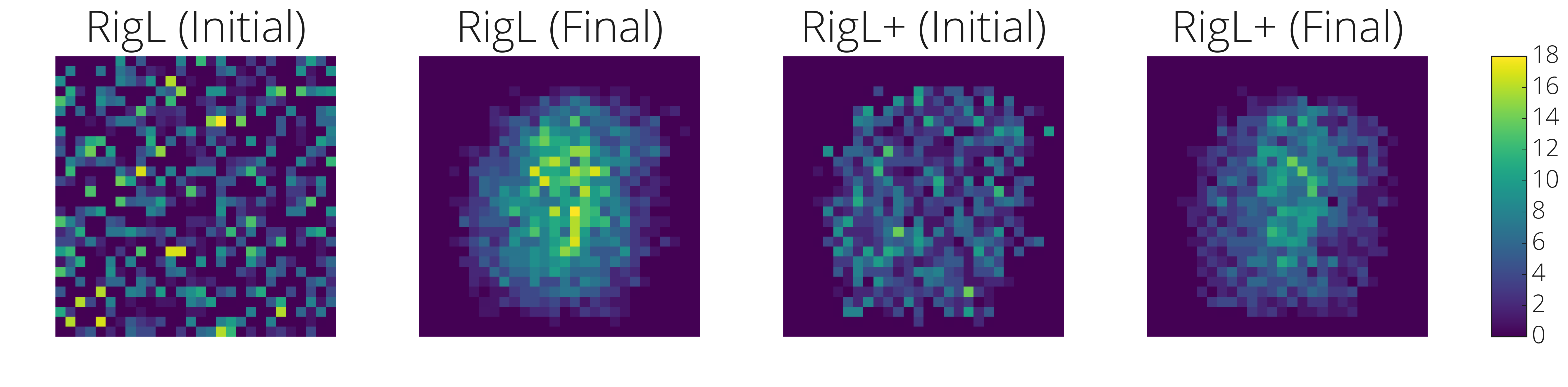}
\caption{Number of connections that originate from the pixels of MNIST images at the beginning and end of the training. RigL+ starts from a smaller architecture (408-100-69) that has already removed some of the input pixels near the edges. Starting from an initially random distribution, \textit{RigL} converges on the most relevant dimensions. See main text for further details.}
\label{fig:app:mnist}
\end{figure*}

To compare, we train a sparse network starting from the original MLP architecture (\textbf{RigL}). At initialization, we randomly remove 99\% and 89\% of the connections in the first and second layer of the MLP.  At the end of the training many of the neurons in the first 2 layers have no in-coming or out-going connections. We remove such neurons and use the resulting architecture to calculate the inference FLOPs and the size. We assume the sparse connectivity is stored as a bit-mask (We assume parameters are represented as floats, i.e. 4 bytes). In this setting, \textit{RigL} finds smaller, more FLOP efficient networks with far less work than the Bayesian approaches.

Next, we train a sparse network starting from the architecture found by the first run (\textbf{RigL+}) (408-100-69) but with a new random initialization (both masks and the parameters). 
We reduce the sparsity of the first 2 layers to 96\% and 86\% respectively as the network is already much smaller. Repeating \textit{RigL} training results in an even more compact architecture half the size and requiring only a third the FLOPs of the best architecture found by~\citet{vib2018}. 

Examination of the open-sourced code for the methods considered here made us aware that all of them keep track of the test error during training and report the best error ever observed during training as the final error. We generally would not encourage such overfitting to the test/validation set, however to make the comparisons with these results fair we report both the lowest error observed during training and the error at the end of training (reported in parenthesis).  All hyper-parameter tuning was done using only the final test error.

\begin{figure*}
\centering
\begin{minipage}{.5\textwidth}
  \centering
  \includegraphics[width=.95\linewidth]{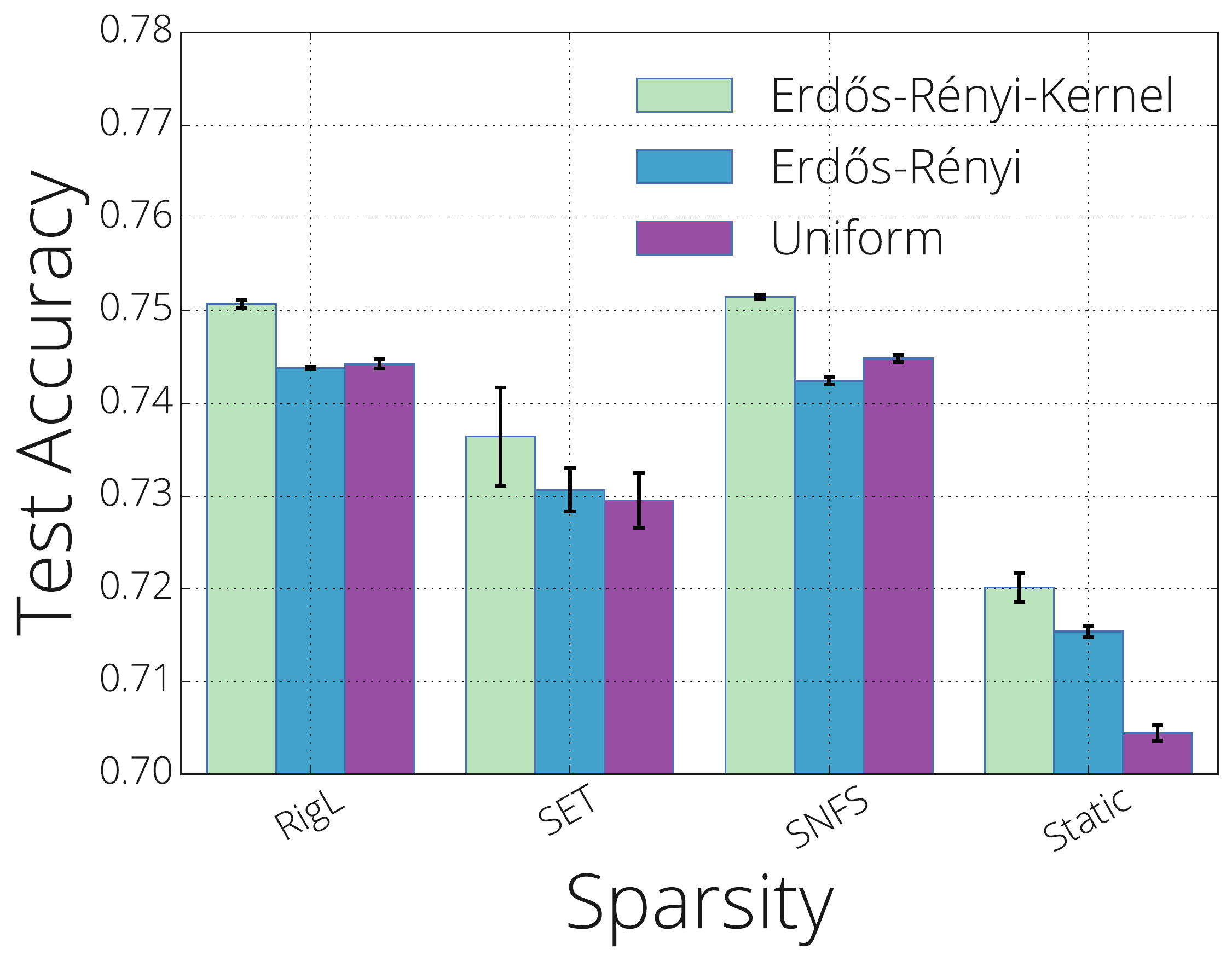}
\end{minipage}%
\begin{minipage}{.5\textwidth}
  \centering
  \includegraphics[width=.9\linewidth]{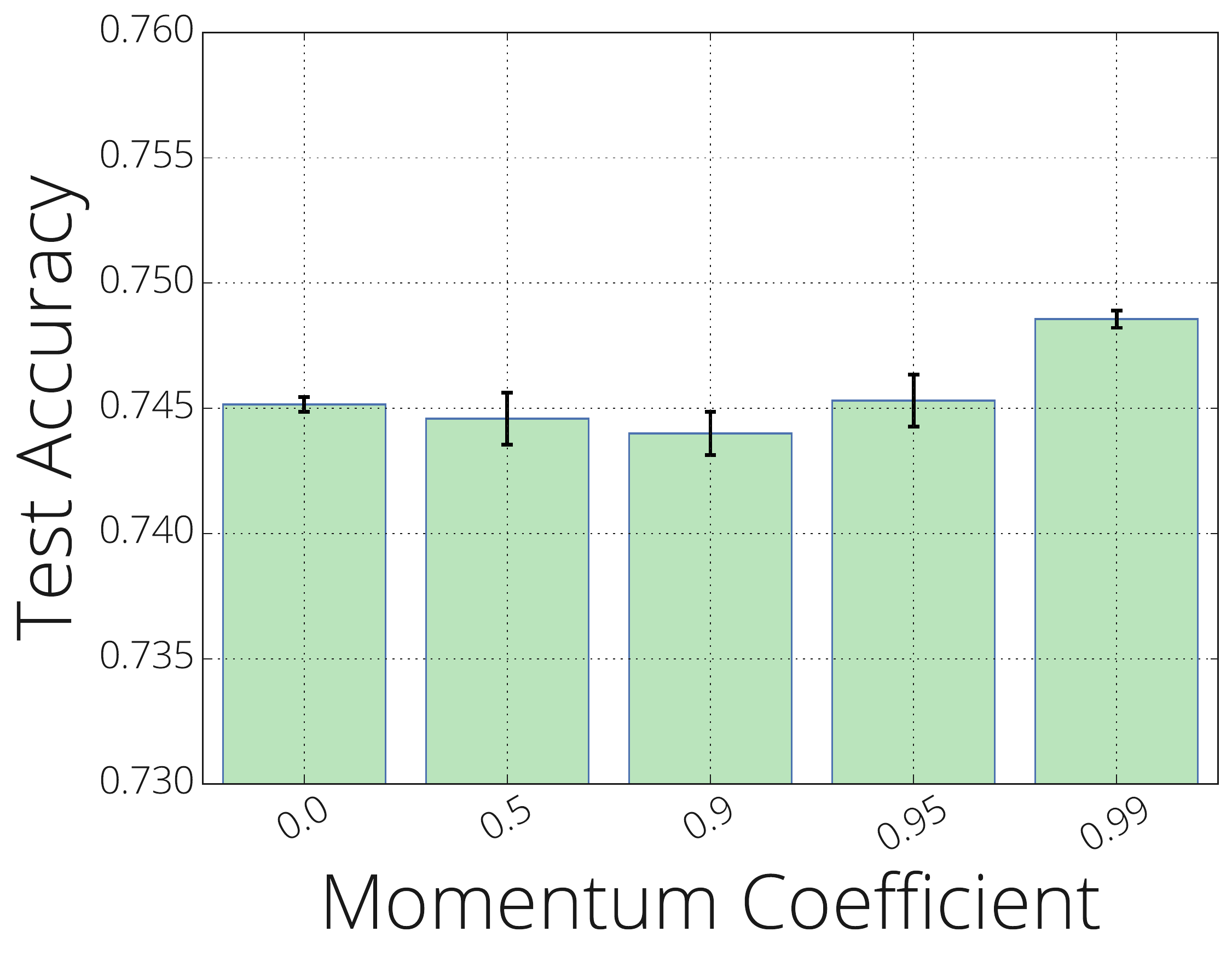}
\end{minipage}
\caption{\textbf{(left)} Effect of sparsity distribution choice on sparse training methods at different sparsity levels. We average over 3 runs and report the standard deviations for each. \textbf{(right)} Effect of momentum value on the performance of SNFS algorithm. Momentum does not become helpful until it reaches extremely large values.}
\label{fig:app:maskinit_and_momentum}
\end{figure*}

In Figure \ref{fig:app:mnist} we visualize how RigL chooses to connect to the input and how this evolves from the beginning to the end of training.  The heatmap shows the number of outgoing connections from each input pixels at the beginning (RigL Initial) and at the end (RigL (Final)) of the training. The left two images are for the initial network and the right two images are for \textbf{RigL+} training.  \textit{RigL} automatically discards uninformative pixels and allocates the connections towards the center highlighting the potential of \textit{RigL} on model compression and feature selection.

\section{Effect of Sparsity Distribution on Other Methods}
\label{app:initmethods}
In Figure \ref{fig:app:maskinit_and_momentum}-left we show the effect of sparsity distribution choice on 4 different sparse training methods. ERK distribution performs better than other distributions for each training method. 

\section{Effect of Momentum Coefficient for SNFS}
\label{app:momentum}
In Figure~\ref{fig:app:maskinit_and_momentum} right we show the effect of the momentum coefficient on the performance of SNFS. Our results shows that using a coefficient of 0.99 brings the best performance. On the other hand using the most recent gradient only (coefficient of 0) performs as good as using a coefficient of 0.9. This result might be due to the large batch size we are using (4096), but it still motivates using {\em RigL} and instantaneous gradient information only when needed, instead of accumulating them.

\section{(Non)-Existence of Lottery Tickets}
\label{app:lottery}
We perform the following experiment to see whether \textit{Lottery Tickets} exist in our setting.  We take the sparse network found by {\em RigL} and restart training using original initialization, both with RigL and with fixed topology as in the original Lottery Ticket Hypothesis.  Results in table~\ref{tab:app:lottery} demonstrate that training with a fixed topology is significantly worse than training with {\em RigL} and that {\em RigL} does not benefit from starting again with the final topology and the original initialization - training for twice as long instead of rewiring is more effective. In short, there are no special tickets, with {\em RigL} all tickets seems to win.

\begin{table*}
    \centering
    \begin{tabular}{lccc} \toprule
        Initialization & Training Method & Test Accuracy & Training FLOPs \\\midrule
        Lottery & Static & 70.82\ci{0.07} & 0.46x \\
        Lottery & RigL & 73.93\ci{0.09} & 0.46x \\
        Random & RigL & 74.55\ci{0.06}& 0.23x \\
        Random & RigL$_{2\times}$ & 76.06\ci{0.09} & 0.46x \\
\bottomrule
    \end{tabular}
    \caption{Effect of lottery ticket initialization on the final performance. There are no special tickets and the dynamic connectivity provided by {\em RigL} is critical for good performance.}
    \label{tab:app:lottery}
\end{table*}

\section{Effect of Update Schedules on Other Dynamic Sparse Methods}
\label{app:schedules}
\begin{figure*}[ht]
\centering
\begin{minipage}{.5\textwidth}
  \centering
  \includegraphics[width=.95\linewidth]{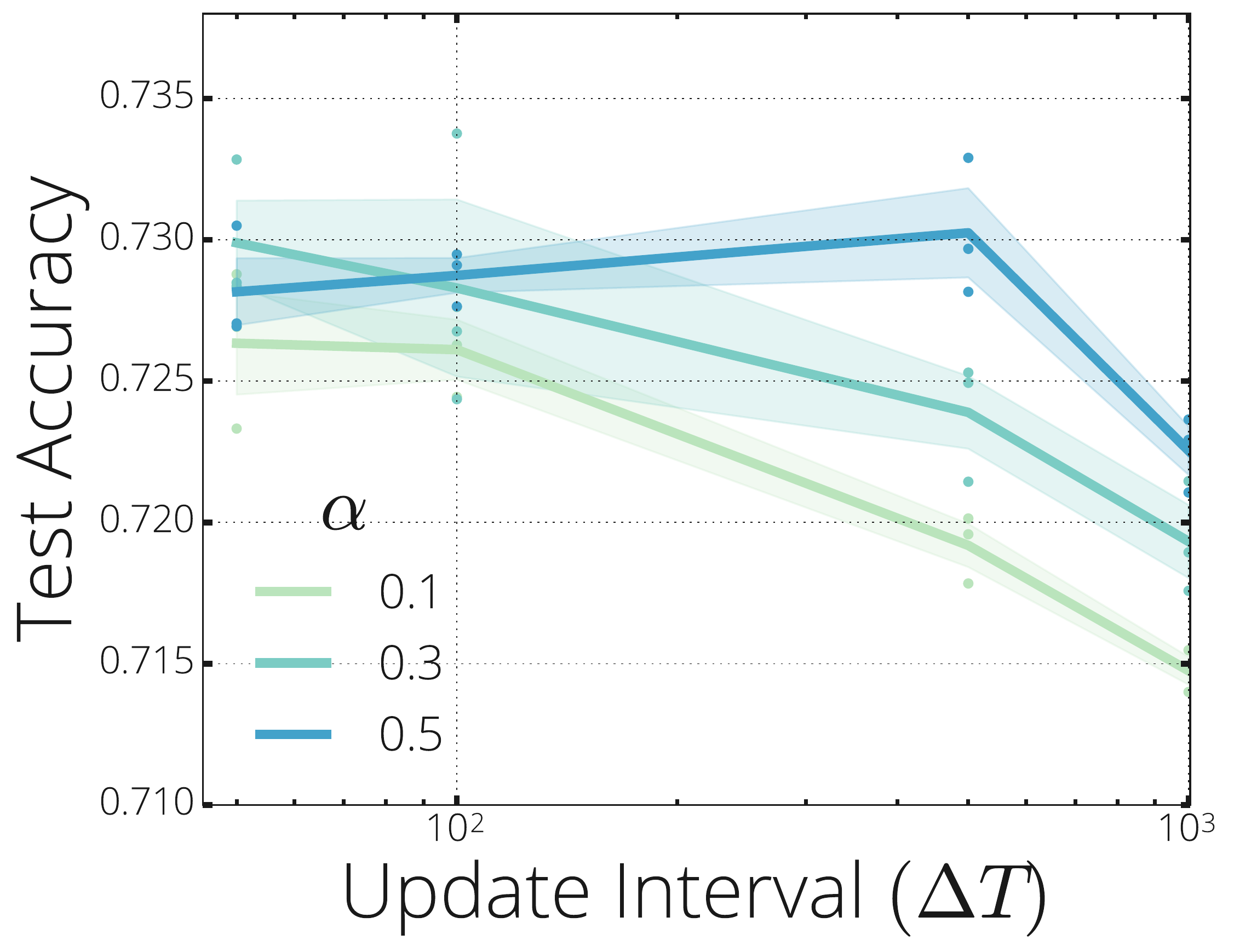}
\end{minipage}%
\begin{minipage}{.5\textwidth}
  \centering
  \includegraphics[width=.9\linewidth]{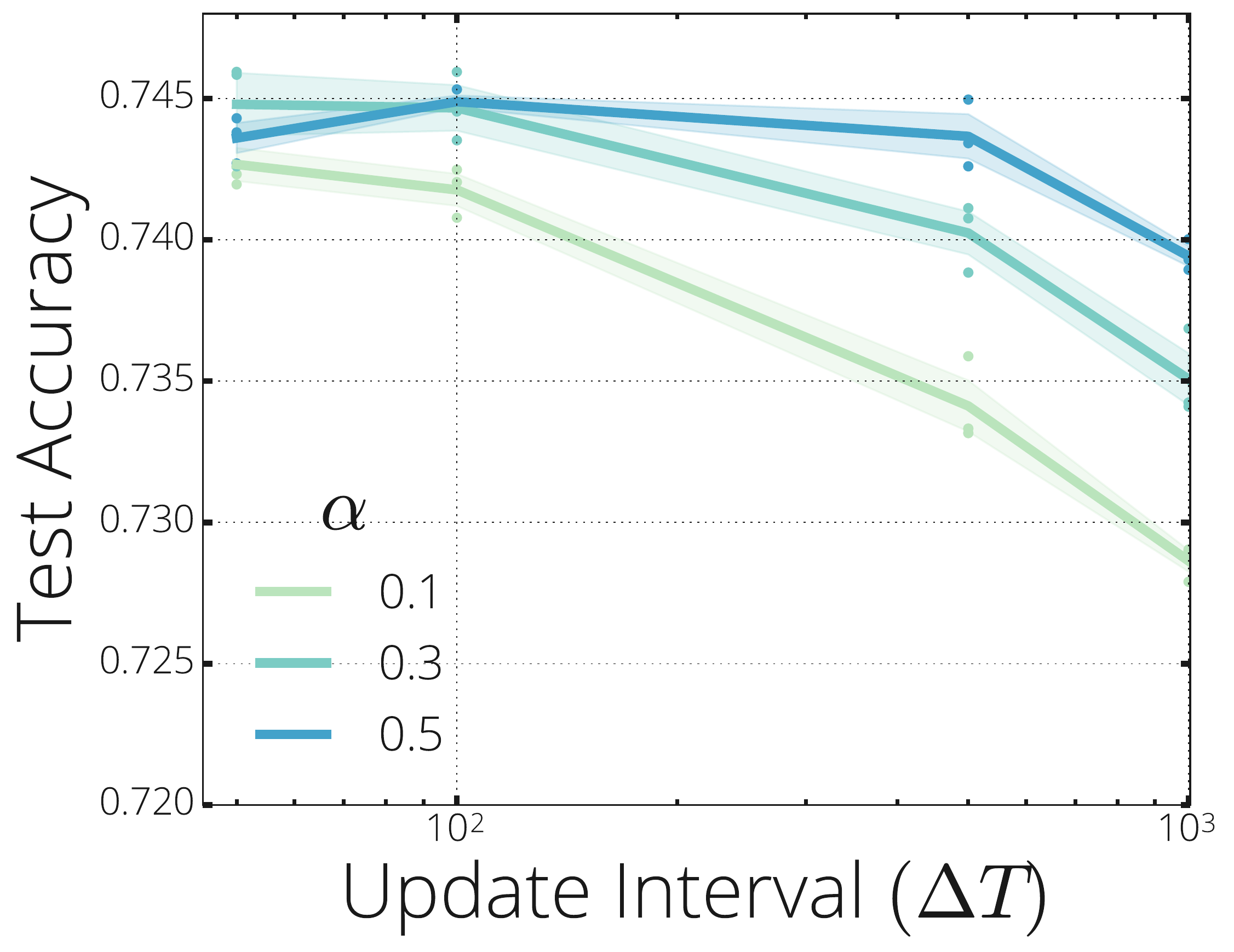}
\end{minipage}
\caption{Cosine update schedule hyper-parameter sweep done using dynamic sparse training methods SET \textbf{(left)} and SNFS \textbf{(right)}.}
\label{fig:app:update_schedule_methods}
\end{figure*}

In \autoref{fig:app:update_schedule_methods} we repeat the hyper-parameter sweep done for {\em RigL} in \autoref{fig:ablation}-right, using SET and SNFS. Cosine schedule with $\Delta T=50$ and $\alpha=0.1$ seems to work best across all methods. An interesting observation is that higher drop fractions ($\alpha$) seem to work better with longer intervals $\Delta T$. For example, SET with $\Delta T=1000$ seems to work best with $\alpha=0.5$.

\section{Alternative Update Schedules}
\label{app:schedules_other}
\begin{figure*}
\centering
\begin{minipage}{.33\textwidth}
  \centering
  \includegraphics[width=.95\linewidth]{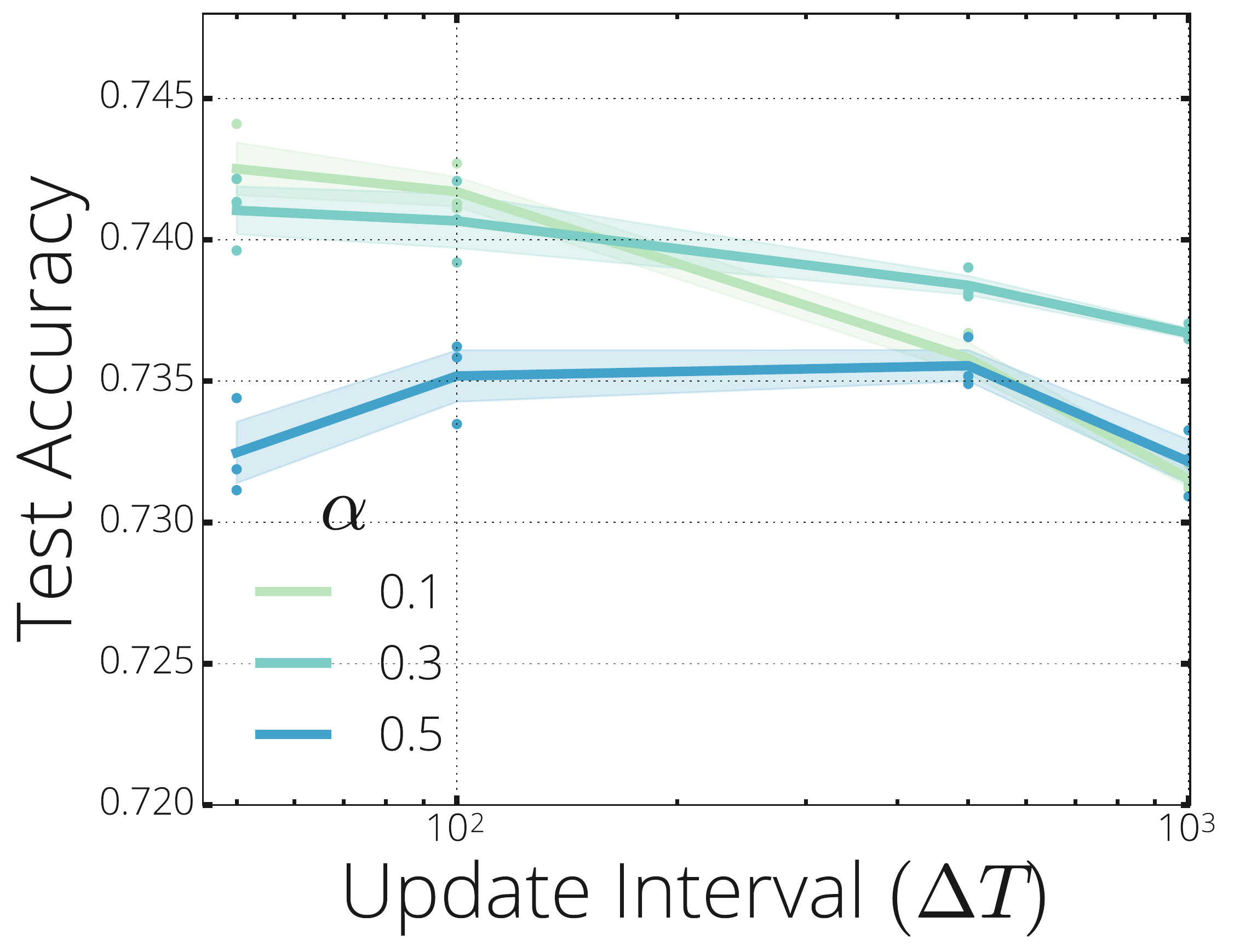}
\end{minipage}%
\begin{minipage}{.33\textwidth}
  \centering
  \includegraphics[width=.9\linewidth]{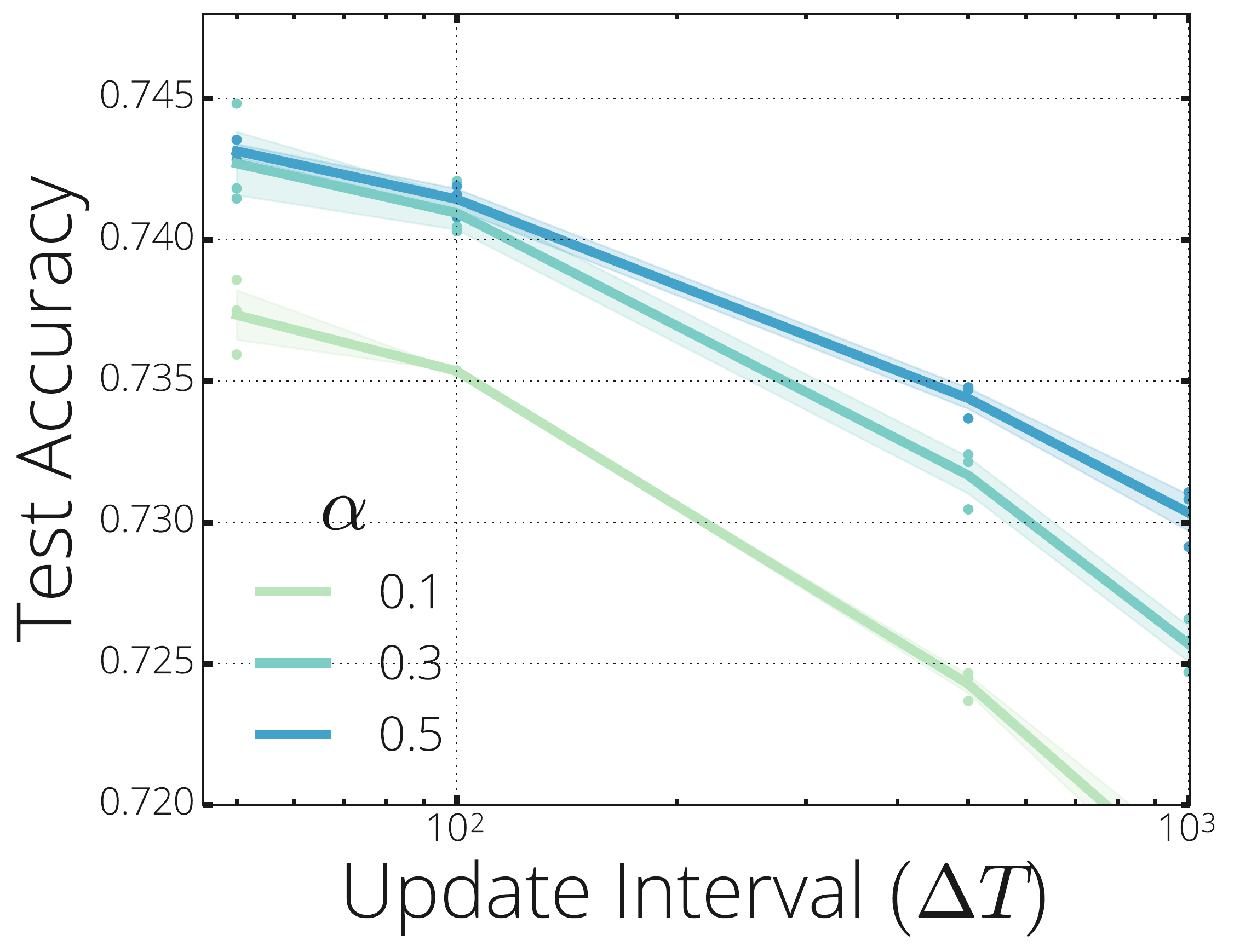}
\end{minipage}
\begin{minipage}{.33\textwidth}
  \centering
  \includegraphics[width=.9\linewidth]{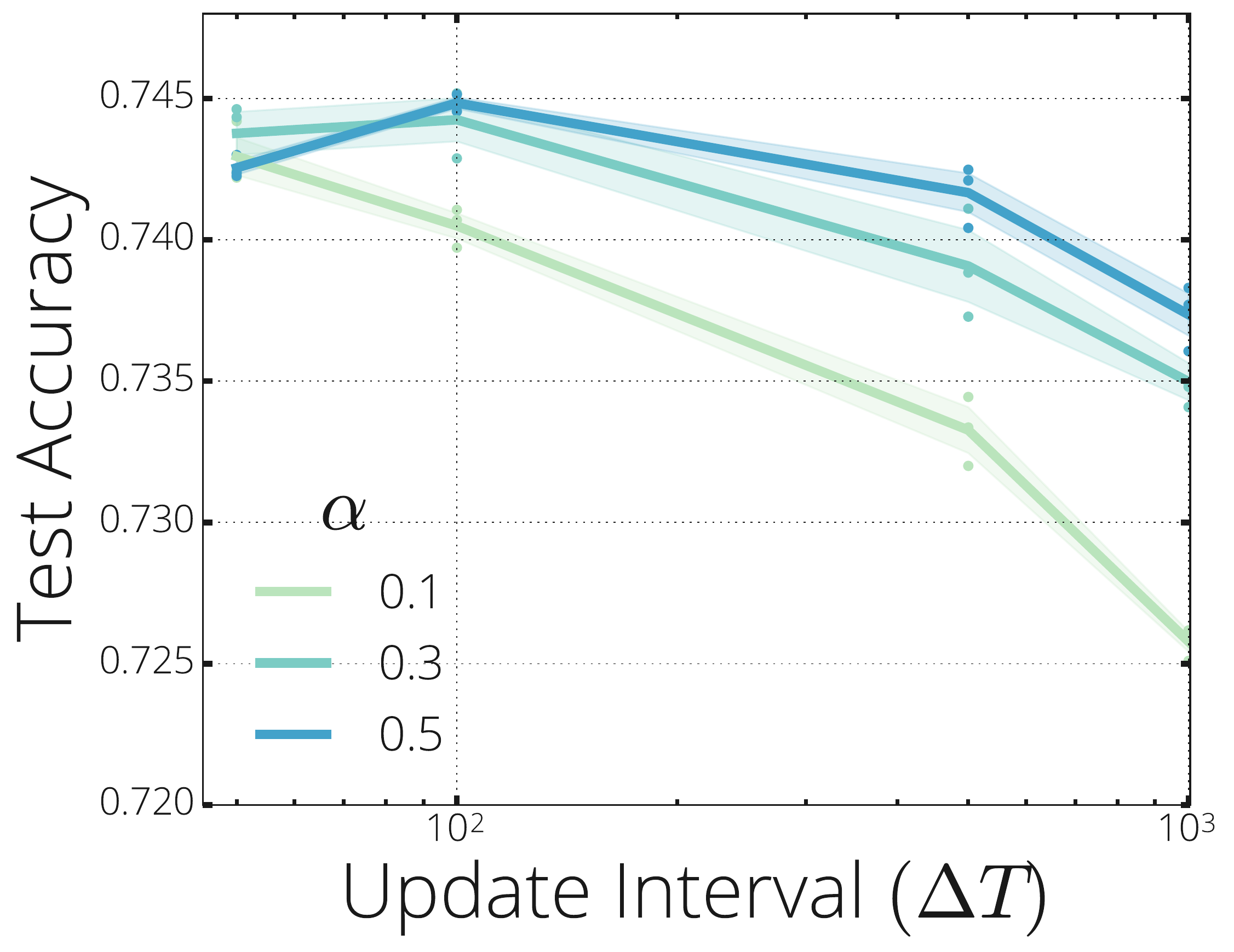}
\end{minipage}
\caption{Using other update schedules with {\em RigL}: \textbf{(left)} Constant \textbf{(middle)} Exponential (k=3) and \textbf{(right)} Linear }
\label{fig:app:schedules_other}
\end{figure*}

In Figure \ref{fig:app:schedules_other}, we share the performance of two alternative annealing functions:
\begin{enumerate}
    \item \textit{Constant:} $f_{decay}(t)=\alpha$.
    \item \textit{Inverse Power:} The fraction of weights updated decreases similarly to the schedule used in~\cite{gupta2018} for iterative pruning: $f_{decay}(t)=\alpha(1-\frac{t}{T_{end}})^k$. In our experiments we tried $k=1$ which is the linear decay and their default $k=3$.
\end{enumerate}
\textit{Constant} seems to perform well with low initial drop fractions like $\alpha=0.1$, but it starts to perform worse with increasing $\alpha$. \textit{Inverse Power} for k=3 and k=1 (\textit{Linear}) seems to perform similarly for low $\alpha$ values. However the performance drops noticeably for k=3 when we increase the update interval. As reported by \cite{dettmers2019} linear (k=1) seems to provide similar results as the cosine schedule.

\section{Calculating FLOPs of models and methods}
\label{app:flops}
In order to calculate FLOPs needed for a single forward pass of a sparse model, we count the total number of multiplications and additions layer by layer for a given layer sparsity $s^l$. The total FLOPs is then obtained by summing up all of these multiply and adds. 
\begin{figure*}[b!]
\centering
\begin{minipage}{.5\textwidth}
  \centering
  \includegraphics[width=.9\linewidth]{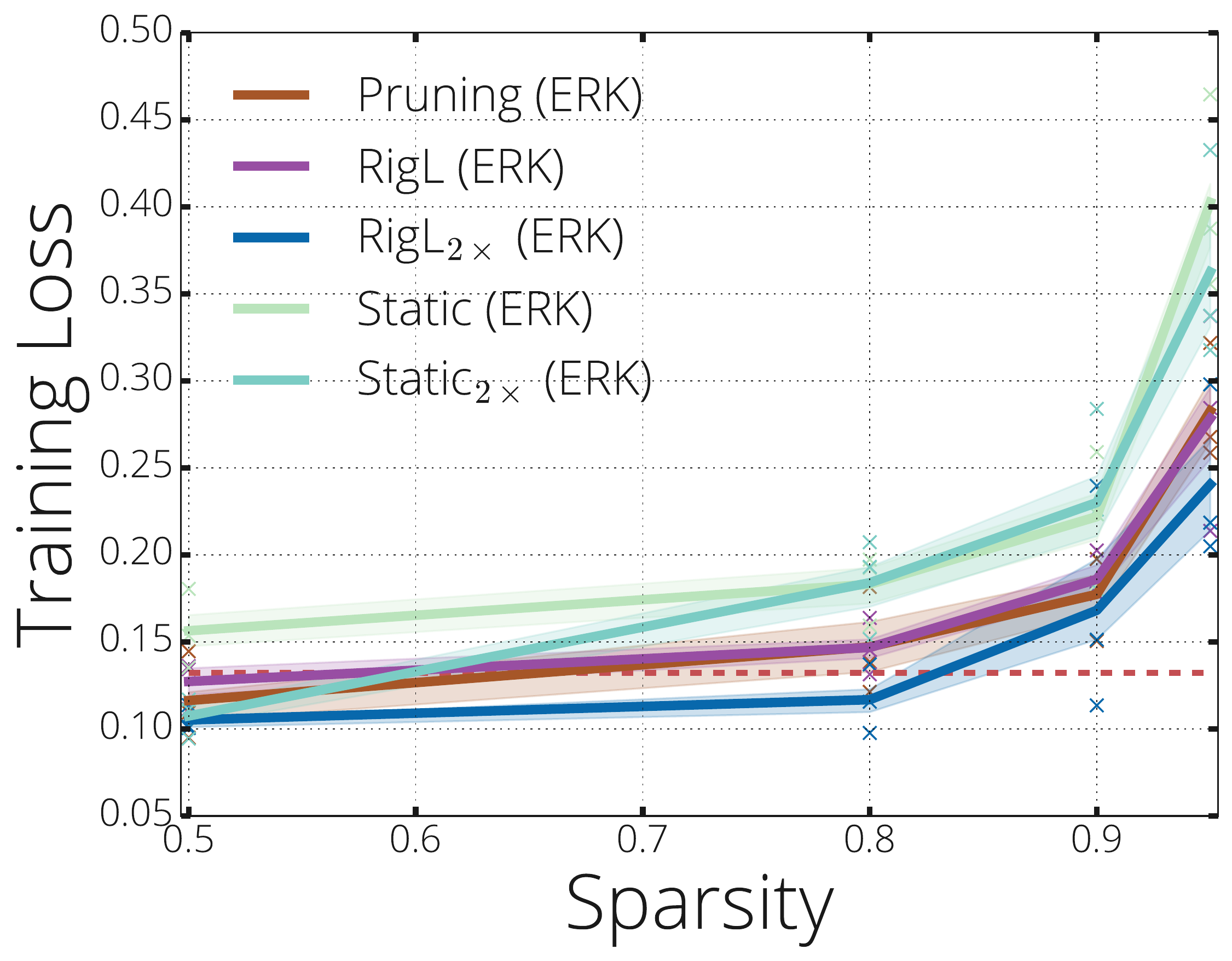}
\end{minipage}%
\begin{minipage}{.5\textwidth}
  \centering
  \includegraphics[width=.9\linewidth]{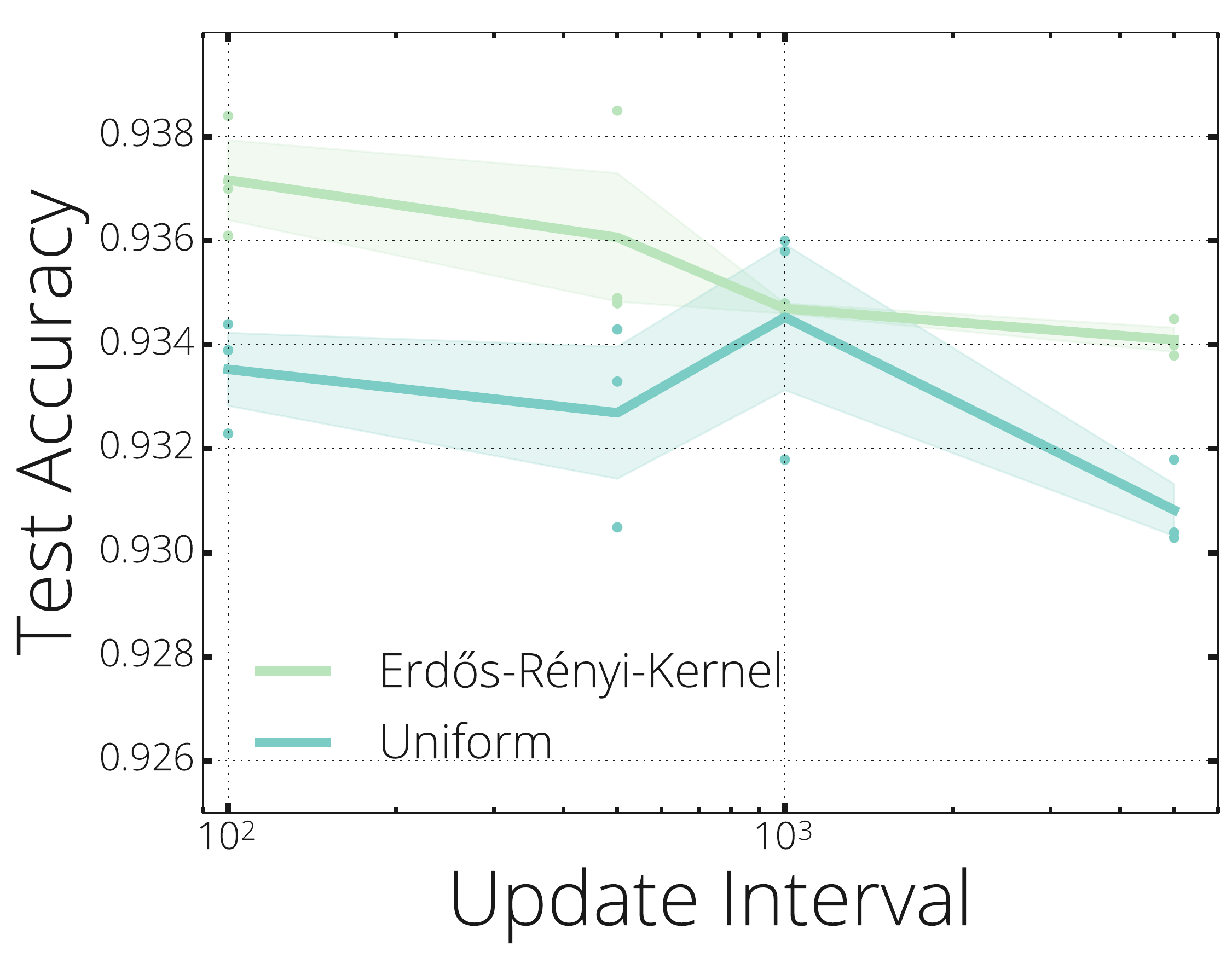}
\end{minipage}
\caption{Final training loss of sparse models \textbf{(left)} and performance of {\em RigL} at different mask update intervals \textbf{(right)}.}
\label{fig:app:cifar10}
\end{figure*}
Different sparsity distributions require different number of FLOPs to compute a single prediction. For example \textit{Erdős-Renyi-Kernel} distributions usually cause 1x1 convolutions to be less sparse than the 3x3 bottleneck layers (see Appendix \ref{app:sparsities}). The number of input/output channels of 1x1 convolutional layers are greater and therefore require more FLOPs to compute the output features compared to 3x3 layers of the ResNet blocks. Thus, allocating smaller sparsities to 1x1 convolutional layers results in a higher overall FLOPs than a sparse network with uniform sparsity.

Training a neural network consists of 2 main steps:
\begin{enumerate}
    \item \textit{forward pass:} Calculating the loss of the current set of parameters on a given batch of data. During this process layer activations are calculated in sequence using the previous activations and the parameters of the layer. Activation of layers are stored in memory for the backward pass.
    \item \textit{backward pass:} Using the loss value as the initial error signal, we back-propagate the error signal while calculating the gradient of parameters. During the backward pass each layer calculates 2 quantities: the gradient of the activations of the previous layer and the gradient of its parameters. Therefore in our calculations we count backward passes as two times the computational expense of the forward pass. We omit the FLOPs needed for batch normalization and cross entropy.
\end{enumerate}

Dynamic sparse training methods require some extra FLOPs to update the connectivity of the neural network. We omit FLOPs needed for dropping the lowest magnitude connections in our calculations. For a given dense architecture with FLOPs $f_D$ and a sparse version with FLOPs $f_S$, the total FLOPs required to calculate the gradient on a single sample is computed as follows:
\begin{itemize}
    \item \textbf{Static Sparse and Dense.} Scales with $3*f_S$ and $3*f_D$ FLOPs, respectively.
    \item \textbf{Pruning.} $\mathbb{E}_t[3*f_D*(1-s_t)$]  FLOPs where $s_t$ is the sparsity of the model at iteration $t$.
    \item \textbf{Snip.} We omit the initial dense gradient calculation since it is negligible, which means Snip scales in the same way as Static methods:  $3*f_S$ FLOPs.
    \item \textbf{SET.} We omit the extra FLOPs needed for growing random connections, since this operation can be done on chip efficiently. Therefore, the total FLOPs for SET scales with $3*f_S$.
    \item \textbf{SNFS.} Forward pass and back-propagating the error signal needs $2*f_S$ FLOPs. However, the dense gradient needs to be calculated at every iteration. Thus, the total number of FLOPs scales with $2*f_S+f_D$.
    \item \textbf{RigL.} Iterations with no connection updates need $3*f_S$ FLOPs. However, at every $\Delta T$ iteration we need to calculate the dense gradients. This results in the average FLOPs for {\em RigL} given by $\frac{(3*f_S*\Delta T+2*f_S+f_D)}{(\Delta T+1)}$.
\end{itemize}

\section{Hyper-parameters used in Charachter Level Language Modeling Experiments}
\label{app:chargru}
As stated in the main text, our network consists of a shared embedding with dimensionality 128, a vocabulary size of 256, a GRU with a state size of 512, a readout from the GRU state consisting of two linear layers with width 256 and 128 respectively.  We train the next step prediction task with the cross entropy loss using the Adam optimizer. We set the learning rate to $7e-4$ and L2 regularization coefficient to $5e-4$. We use a sequence length of 512 and a batch size of 32. Gradients are clipped when their magnitudes exceed 10. We set the sparsity to 75\% for all models and run 200,000 iterations.  When inducing sparsity with magnitude pruning~\citep{gupta2018}, we perform pruning between iterations 50,000 and 150,000 with a frequency of 1,000. We initialize sparse networks with a uniform sparsity distribution and use a cosine update schedule with $\alpha=0.1$ and $\Delta T=100$. Unlike the previous experiments we keep updating the mask until the end of the training since we observed this performed slightly better than stopping at iteration 150,000.

\section{Additional Plots and Experiments for CIFAR-10}
\label{app:cifar10}
In Figure \ref{fig:app:cifar10}-left, we plot the final training loss of experiments presented in Section \ref{sec:cifar10} to investigate the generalization properties of the algorithms considered. Poor performance of \textit{Static} reflects itself in training loss clearly across all sparsity levels. \textit{RigL} achieves similar final loss as the pruning, despite having around half percent less accuracy. Training longer with \textit{RigL} decreases the final loss further and the test accuracies start matching pruning (see Figure \ref{fig:wikichar_cifar10}-right) performance. These results show that \textit{RigL} improves the optimization as promised, however generalizes slightly worse than pruning.  

In Figure \ref{fig:app:cifar10}-right, we sweep mask update interval $\Delta T$ and plot the final test accuracies. We fix initial drop fraction $\alpha$ to $0.3$ and evaluate two different sparsity distributions: \textit{Uniform} and \textit{ERK}. Both curves follow a similar pattern as in Imagenet-2012 sweeps (see Figure \ref{fig:app:update_schedule_methods}) and best results are obtained when $\Delta T=100$.

\section{Sparsity of Individual Layers for Sparse ResNet-50}
\label{app:sparsities}
\begin{figure*}
\centering
  \includegraphics[width=.8\linewidth]{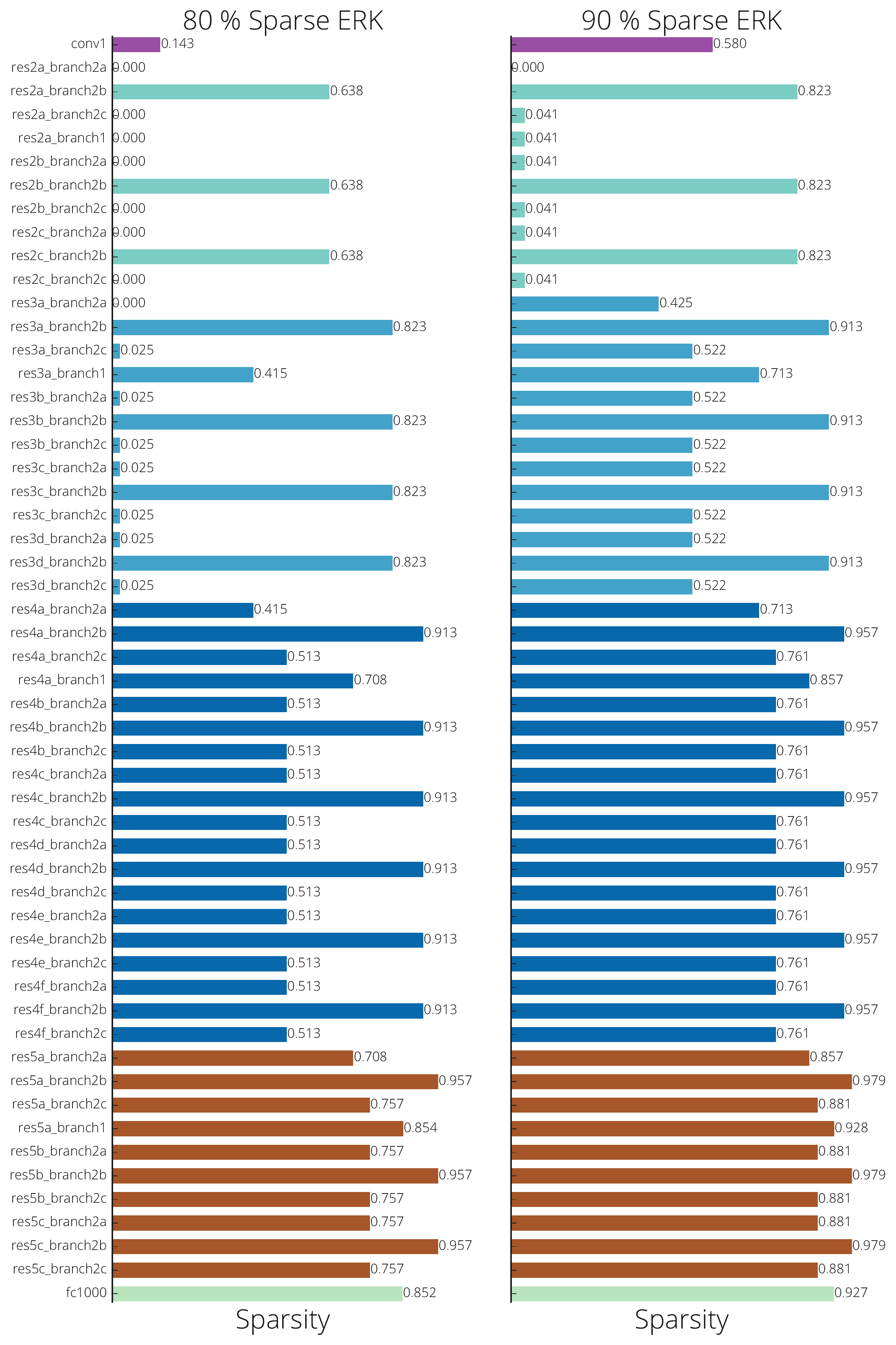}
\caption{Sparsities of individual layers of the ResNet-50.}
\label{fig:app:sparsities}
\end{figure*}

Sparsity of ResNet-50 layers given by the Erdős-Rényi-Kernel sparsity distribution plotted in Figure \ref{fig:app:sparsities}. 

\section{Performance of Algortihms at Training 95 and 96.5\% Sparse ResNet-50}
In this section we share results of algorithms at training ResNet-50s with higher sparsities. Results in Table \ref{fig:app:highsparsity} indicate \textit{RigL} achieves higher performance than the pruning algorithm even without extending training length. 
\begin{table*}
\centering
\begin{tabular}{c|p{5em}|p{4em}|p{4em}||p{5em}|p{4em}|p{4em}}
    \toprule
     Method & Top-1 \newline Accuracy& FLOPs \newline (Train) & FLOPs\newline (Test) & Top-1 \newline Accuracy& FLOPs \newline (Train) & FLOPs\newline (Test) \\\toprule
    Dense & 76.8\ci{0.09} & 1x \small{(3.2e18)} & 1x \small{(8.2e9)} \\\hline
        & \multicolumn{3}{c ||}{S=0.95} & \multicolumn{3}{ c }{S=0.965}\\\hline \hline
    Static & 59.5+-0.11 & 0.23x & 0.08x & 55.4+-0.06 & 0.13x & 0.07x \\
    Snip & 57.8+-0.40 & 0.23x & 0.08x & 52.0+-0.20 & 0.13x & 0.07x \\
    SET & 64.4+-0.77 & 0.23x & 0.08x & 60.8+-0.45 & 0.13x & 0.07x \\
    RigL & 67.5+-0.10 & 0.23x & 0.08x & 65.0+-0.28 & 0.13x & 0.07x \\
    RigL$_{5\times}$ & 73.1+-0.12 & 1.14x & 0.08x & 71.1+-0.20 & 0.66x & 0.07x \\\hline
    Static (ERK) & 72.1\ci{0.04} & 0.42x & 0.42x & 67.7\ci{0.12} & 0.24x & 0.24x \\
    RigL (ERK) & 69.7+-0.17 & 0.42x & 0.12x & 67.2+-0.06 & 0.25x & 0.11x \\
RigL$_{5\times}$ (ERK) & 74.5+-0.09 & 2.09x & 0.12x & 72.7+-0.02 & 1.23x & 0.11x \\\hline \hline
    SNFS (ERK) & 70.0+-0.04 & 0.61x & 0.12x & 67.1+-0.72 & 0.50x & 0.11x \\
    Pruning* (Gale) & 70.6 & 0.56x & 0.08x & n/a & 0.51x & 0.07x \\
    Pruning$_{1.5\times}$ (Gale) & 72.7 & 0.84x & 0.08x & 69.26 & 0.76x & 0.07x 
    \\
    \bottomrule
    \end{tabular}
\caption{Results with increased sparsity on ResNet-50/ImageNet-2012.}
\label{fig:app:highsparsity}
\end{table*}

\section{Bugs Discovered During Experiments}
Our initial implementations contained some subtle bugs, which while not affecting the general conclusion that {\em RigL} is more effective than other techniques, did result in lower accuracy for all sparse training techniques.  We detail these issues here with the hope that others may learn from our mistakes.

\begin{enumerate}
    \item \textbf{Random operations on multiple replicas.}  We use data parallelism to split a mini-batch among multiple replicas.  Each replica independently calculates the gradients using a different sub-mini-batch of data. The gradients are aggregated using an \textsc{all-reduce} operation before the optimizer update. Our implementation of SET, SNFS and {\em RigL} depended on each replica independently choosing to drop and grow the same connections.  However, due to the nature of random operations in Tensorflow, this did not happen.  Instead, different replicas diverged after the first drop/grow step.  This was most pronounced in SET where each replica chose at random and much less so for SNFS and {\em RigL} where randomness is only needed to break ties. If left unchecked this might be expected to be catastrophic, but due to the behavior of Estimators and/or TF-replicator, the values on the first replica are broadcast to the others periodically (every approximately 1000 steps in our case).
    
     We fixed this bug by using \href{https://www.tensorflow.org/api_docs/python/tf/random/stateless_uniform}{stateless random operations}. As a result the performance of SET improved slightly (0.1-0.3 \% higher on Figure \ref{fig:resnet}-left).
    
    \item \textbf{Synchronization between replicas.} {\em RigL} and SNFS depend on calculating dense gradients with respect to the masked parameters. However, as explained above, in the multiple replica setting these gradients need to be aggregated. Normally this aggregation is automatically done by the optimizer, but in our case, this does not happen (only the gradients with respect to the \emph{unmasked} parameters are aggregated automatically).  This bug affected SNFS and {\em RigL}, but not SET since SET does not rely on the gradients to grow connections.  Again, the synchronization of the parameters from the first replica every approximately 1000 steps masked this bug.
    
    We fixed this bug by explicitly calling \textsc{all-reduce} on the gradients with respect to the masked parameters. With this fix, the performance of {\em RigL} and SNFS improved significantly, particularly for default training lengths (around 0.5-1\% improvement).
    
    \item \textbf{SNIP Experiments.} Our first implementation of SNIP used the gradient magnitudes to decide which connections to keep causing its performance to be worse than static. Upon our discussions with the authors of SNIP, we realized that the correct metric is the saliency (gradient times parameter magnitude). With this correction SNIP performance improved dramatically to better than random (Static) even at Resnet-50/ImageNet scale. It is surprising that picking connections with the highest gradient magnitudes can be so detrimental to training (it resulted in much worse than random performance).
\end{enumerate}

\section{Changes Since Publishing}
\begin{enumerate}
    \item The ordering of SET and NeST updated after learning SET was published earlier
    \item $\mathbb{I}_{drop}$ notation is replaced by $\mathbb{I}_{active}$ to make it more clear that the indices captured by the notation are the connections kept.
    \item Typo at pruning flops calculation is fixed (kudos to Varun Sundar for reporting). 
    \item Zero initialization at growth is emphasized and together with some other choices attempted.
\end{enumerate}

\end{document}